\documentclass[10pt,twocolumn,letterpaper]{article}

\usepackage{cvpr}              

\usepackage{multirow}

\usepackage{amssymb}   
\usepackage{array}
\newcommand{\cmark}{\ensuremath{\checkmark}}
\newcommand{\xmark}{\ensuremath{\times}}
\newcommand{\lmark}{\ensuremath{\sim}}
\DeclareUnicodeCharacter{2194}{\leftrightarrow}

\definecolor{cvprblue}{rgb}{0.21,0.49,0.74}
\usepackage[pagebackref,breaklinks,colorlinks,allcolors=cvprblue]{hyperref}

\def\confName{CVPR}
\def\confYear{2026}

\title{Learning Latent Proxies for Controllable Single-Image Relighting}

\author{%
  Haoze Zheng$^{1*}$,
  Zihao Wang$^{1*}$,
  Xianfeng Wu$^{1*}$,
  Yajing Bai$^{1}$,
  Yexin Liu$^{1}$,\\ 
  Yun Li$^{2}$,
  Xiaogang Xu$^{3\dagger}$,
  Harry Yang$^{1\dagger}$\\[0.5em]
  $^1$HKUST \quad
  $^2$HKPolyU \quad
  $^3$CUHK\\
  \small{$^*$Equal contribution \quad $^\dagger$Corresponding authors}\\[0.5em]
  \tt\small \{hzhengay, yangharry\}@connect.ust.hk, xiaogangxu00@gmail.com 
}

\begin{document}

\twocolumn[{%
    \renewcommand\twocolumn[1][]{#1}%
    \maketitle
    \begin{center}
    \vspace{-30pt}
    \includegraphics[width=1.0\textwidth]{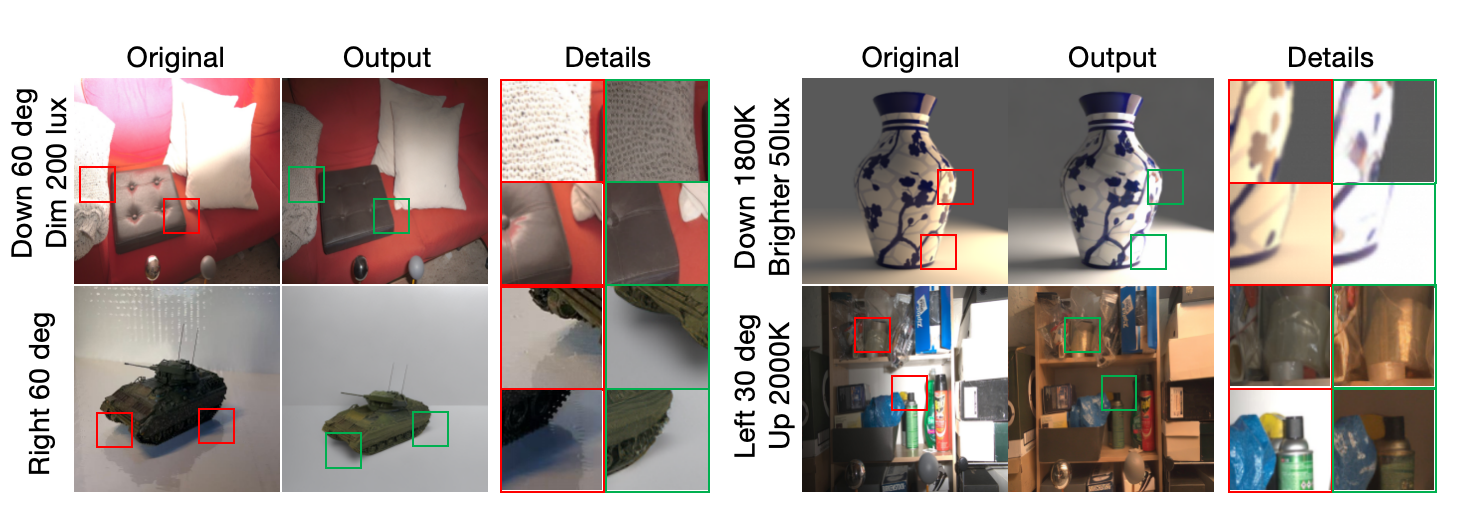}
    \vspace{-30pt}
    \captionsetup{hypcap=false}
    \captionof{figure}{Our method enables \emph{precise and continuous control} over illumination, 
including light \textbf{direction} (e.g., $60^\circ$ down or right), 
\textbf{intensity} (e.g., dimmed by 200 lux or brightened by 50 lux), 
and \textbf{color temperature} (e.g., shifted to 1800K or +2000K). 
Given only a single source image, the model produces physically consistent relighting while preserving fine textures, specular highlights, and material appearance. 
The examples illustrate how directional, photometric, and chromatic adjustments are faithfully reflected in the output. (Note: The bottom-left image was captured with a mobile phone and cropped before processing.)
}
    \label{fig:teaser}
    \end{center}%
}]

\maketitle
\begin{abstract}
Single-image relighting is highly under-constrained: small illumination changes can produce large, nonlinear variations in shading, shadows, and specularities, while geometry and materials remain unobserved. Existing diffusion-based approaches either rely on intrinsic or G-buffer pipelines that require dense and fragile supervision, or operate purely in latent space without physical grounding, making fine-grained control of direction, intensity, and color unreliable. 
We observe that a full intrinsic decomposition is unnecessary and redundant for accurate relighting. Instead, sparse but physically meaningful cues, indicating where illumination should change and how materials should respond, are sufficient to guide a diffusion model. Based on this insight, we introduce \textbf{LightCtrl} that integrates physical priors at two levels: a few-shot latent proxy encoder that extracts compact material-geometry cues from limited PBR supervision, and a lighting-aware mask that identifies sensitive illumination regions and steers the denoiser toward shading relevant pixels. To compensate for scarce PBR data, we refine the proxy branch using a DPO-based objective that enforces physical consistency in the predicted cues.
We also present \textbf{ScaLight}, a large scale object-level dataset with systematically varied illumination and complete cameralight metadata, enabling physically consistent and controllable training. Across object and scene level benchmarks, our method achieves photometrically faithful relighting with accurate continuous control, surpassing prior diffusion and intrinsic-based baselines, including gains of up to \textbf{+2.4 dB} PSNR and \textbf{35\%} lower RMSE under controlled lighting shifts.
\end{abstract}    
\section{Introduction}

Relighting a single image under novel illumination is fundamentally ill-posed: shadows, specularities, and diffuse shading depend on geometry and materials that are not observable from a single RGB input. Small changes in lighting direction, intensity, or color can therefore induce large, nonlinear variations in appearance, while naïve generative models often hallucinate geometry or drift in surface color. These challenges make high quality relighting require not only visual plausibility, but also fine-grained \emph{controllability} and \emph{physical consistency}.

A number of recent works attempt to introduce controllability into diffusion-based relighting, but each addresses only part of the problem. IC-Light~\cite{zhang2025scaling} fine-tunes a pretrained diffusion model for light-conditioned generation and performs well on portraits; however, its limited physical modeling limits generalization to complex scenes and diverse materials. LBM~\cite{chadebec2025lbm} interpolates between lighting conditions in latent space, producing smooth transitions but offering little physical grounding and weak disentanglement of direction or intensity. LumiNet~\cite{xing2025luminetlatentintrinsicsmeets} embeds relighting into a latent-intrinsic representation, which improves scene level transfer but trades off interpretability and struggles with precise control over color and direction. Neural LightRig~\cite{he2024neurallightrigunlockingaccurate} incorporates physical priors via a multi-stage G-buffer pipeline, yet its dependence on dense PBR supervision makes it fragile and expensive to scale. These methods reveal a persistent gap: strong diffusion priors alone are insufficient for fine-grained lighting control, while physically grounded pipelines require heavy supervision.

These limitations motivate our approach: instead of pursuing full intrinsic decomposition or relying solely on latent diffusion, we aim for a middle ground that retains physical meaningfulness without heavy supervision. Our key observation is that precise relighting does not require complete G-buffers; rather, sparse and spatially targeted physical cues are sufficient to constrain a diffusion model. This motivates our design of a lightweight latent proxy encoder, a lighting-aware mask, and a DPO-refined proxy branch, which collectively provide fine-grained and physically consistent illumination control with markedly lower annotation cost.

We present \textbf{LightCtrl}, a diffusion-based relighting framework that unifies scalable generative modeling with lightweight physical guidance. Our key insight is that precise illumination manipulation does not require full intrinsic reconstruction nor unconstrained latent diffusion; instead, injecting a minimal set of physically meaningful cues into the generative backbone is sufficient to achieve stable and controllable relighting. To this end, we first construct a large scale object level dataset, \textbf{ScaLight}, using a simple yet efficient rendering pipeline that systematically varies lighting direction, intensity, and color while maintaining consistent geometry and materials. We fully fine-tune a Stable Diffusion-based~\cite{meng2022sdeditguidedimagesynthesis} backbone on this dataset to learn generalizable light transport priors. On top of this backbone, we introduce an implicit PBR Encoder, a lightweight few-shot module that predicts a compact latent proxy of material and geometric cues, providing just enough structure to guide illumination changes without requiring dense G-buffer supervision. To further enhance spatial selectivity during denoising, we incorporate a lighting-aware mask module that modulates attention on illumination-sensitive regions across timesteps. Finally, to mitigate the scarcity of PBR-supervised samples and ensure robustness under extreme lighting manipulations, we apply DPO-based~\cite{rafailov2024directpreferenceoptimizationlanguage} post training to the PBR Encoder, significantly improving its physical consistency. By combining these modules, LightCtrl enables precise and robust illumination editing that preserves material appearance and geometry across a broad range of inputs.

In summary, our main contributions are as follows:

\begin{itemize}

  \item We propose \textbf{LightCtrl}, a diffusion-based relighting framework that integrates minimal but physically meaningful priors via a few-shot latent proxy, a lighting-aware mask, and DPO refinement to achieve fine-grained and physically consistent illumination control without requiring dense intrinsic supervision.

  \item We construct \textbf{ScaLight}, a large scale object-level dataset with systematically varied illumination and complete camera–light metadata, enabling controllable, physically consistent training and serving as a comprehensive benchmark for future relighting research.

  \item We demonstrate substantial improvements over intrinsic-based and diffusion baselines on both object and scene-level benchmarks, achieving state-of-the-art RMSE and PSNR as well as the highest user-study preference rate, especially under fine-grained changes in lighting direction, intensity, and color temperature.

\end{itemize}
\section{Related work}

\begin{figure*}[t]
  \centering
  \includegraphics[width=0.7\textwidth]{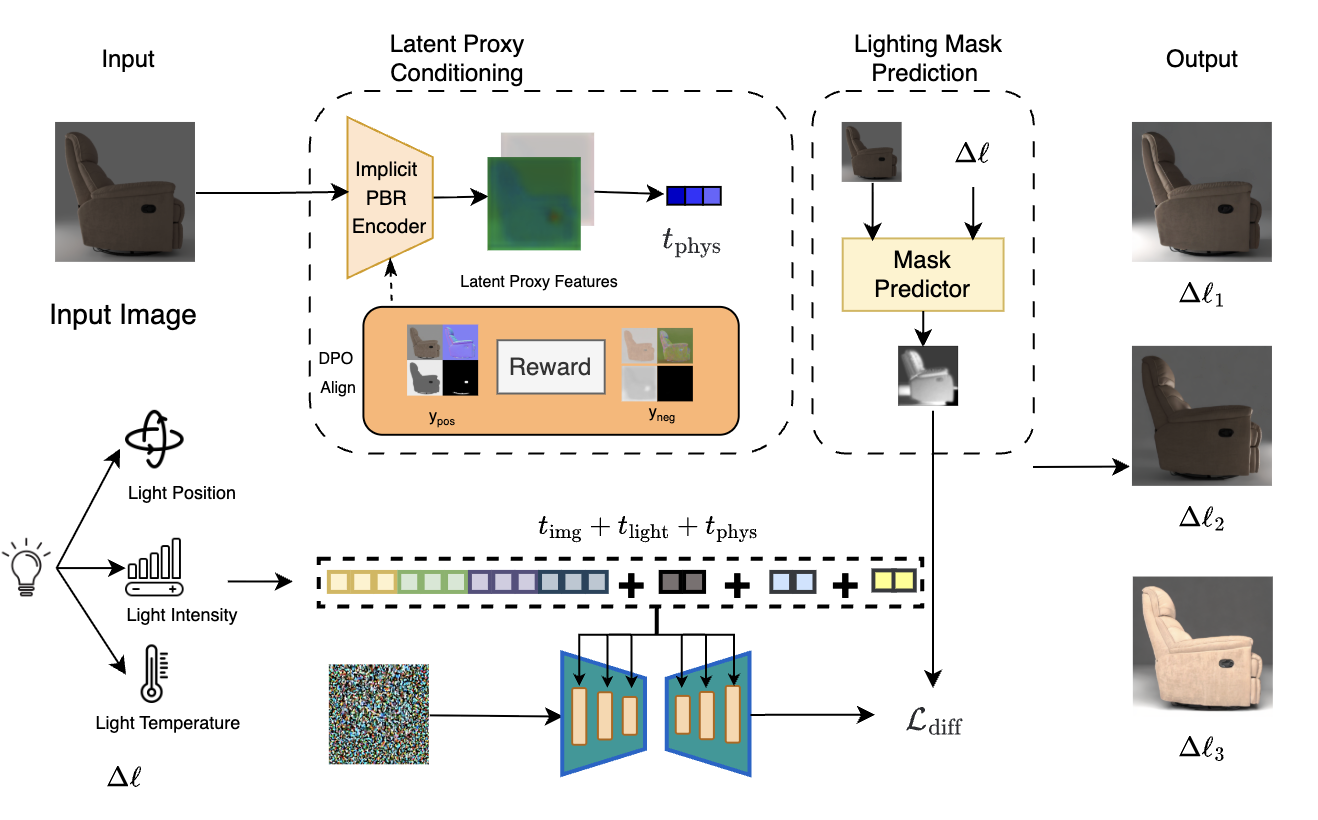}
  \caption{
    Overview of LightCtrl and the object-level rendering pipeline. Our
   model integrates a few-shot latent proxy, a lighting-aware mask, and a
  DPO-refined PBR encoder to condition the UNet during denoising. The implicit
  PBR encoder extracts physically-based latent proxy features $t_{\mathrm{phys}}$, which
  are concatenated with image tokens $t_{\mathrm{image}}$ and light condition tokens
  $t_{\mathrm{light}}$ to guide the diffusion process. A lighting mask predictor takes the lighting change $\Delta\ell$ as input to produce a spatially-aware attention mask for
  the UNet. The output column shows relit results under three different lighting configurations ($\Delta\ell_1$, $\Delta\ell_2$, $\Delta\ell_3$), demonstrating the model's ability to produce diverse and physically plausible relighting effects from a single input image.
  }
  \label{fig:dataoverview}
\end{figure*}

\noindent \textbf{Physics and Intrinsic Guided Relighting.} Early relighting methods relied on inverse rendering and intrinsic decomposition, explicitly estimating geometry, albedo, and illumination to reconstruct physically consistent lighting~\cite{barron2020shapeilluminationreflectanceshading,lin2025geometric}.
While physically interpretable, these methods required strong supervision and often failed to generalize under complex, high-frequency lighting conditions.
With the rise of diffusion~\cite{ho2020denoisingdiffusionprobabilisticmodels, rombach2022highresolutionimagesynthesislatent, song2022denoisingdiffusionimplicitmodels, lu2022dpmsolverfastodesolver, du2025superpc, liu2025unilumosfastunifiedimage} and foundation models~\cite{wu2025qwenimagetechnicalreport}, recent works have shown remarkable capability in modeling illumination and appearance in a data-driven manner.
Zeng et al.~\cite{zeng2024rgb} proposed RGB$\leftrightarrow$X, a two-stage framework that disentangles intrinsic scene representations from RGB inputs to enable realistic and controllable relighting, while Neural Gaffer~\cite{jin2024neuralgafferrelightingobject} further integrates explicit light transport modeling into a diffusion process, demonstrating the benefits of physics-guided constraints in achieving realistic and various illumination control. IllumiNeRF~\cite{zhao2024illuminerf} introduces a 3D relighting framework that combines lighting-conditioned diffusion with NeRF~\cite{mildenhall2020nerfrepresentingscenesneural} reconstruction, circumventing explicit inverse rendering. Unlike the previous methods, it leverages diffusion-generated relit views to build a lighting-aware NeRF for novel-view relighting. Careaga etc.~\cite{Careaga_2025} present a self-supervised relighting framework that enables explicit, physically-based control~\cite{Magar_2025} over light sources by integrating differentiable rendering with neural networks trained on real photographs. These methods demonstrate the effectiveness of explicit modeling, yet their reliance on physical features motivates a shift toward implicit, end-to-end illumination learning.

\noindent \textbf{Implicit and End-to-End Relighting}
Leveraging powerful generative models~\cite{wu2025qwenimagetechnicalreport, labs2025flux1kontextflowmatching, xu2023conditional,wang2025diffdoctor,peebles2023scalablediffusionmodelstransformers}  to synthesize large-scale, illumination-diverse data can significantly enhance a model's ability to learn light control in a self-supervised manner. Building on this idea, IC-Light~\cite{zhang2025scaling} employs a diffusion-based framework trained on a massive synthetic dataset, achieving impressive relighting quality and generalization without relying on intrinsic decomposition. Similar data-driven relighting strategies in recent works~\cite{erel2025practilightpracticallightcontrol, mei2023lightpainterinteractiveportraitrelighting, xu2024low,xu2025learnable,murmann19, zhou2025lightavideotrainingfreevideorelighting} further indicate that high-quality, illumination-diverse datasets are crucial for this task. While these data-driven diffusion methods rely on large-scale synthetic supervision, more recent works shift toward implicit modeling to learn illumination control directly from image features. LumiNet~\cite{xing2025luminetlatentintrinsicsmeets} follows this direction by introducing a latent intrinsic representation that enables continuous and fine-grained relighting under indoor scenarios. Similarly, Ren etc.~\cite{ren2025mvcolightefficientobjectcompositing} formulates relighting as a joint optimization problem in a unified latent space. However, these methods still face limitations, large-scale approaches often require complex data construction pipelines, while simpler latent frameworks lack precise controllability over lighting attributes such as intensity, color, and direction. Our method mitigates the redundancy of two-stage or intrinsic-based frameworks by adopting a lightweight, few-shot training paradigm. It achieves superior light control with a much simpler data construction process, surpassing state-of-the-art methods in both efficiency and precision.

\section{Method}

Given a source image \(x_s^{\,\ell_s}\) and a reference image \(x_r^{\,\ell_t}\) captured under \emph{different} illumination conditions, 
we compute a \emph{relative lighting} representation \(\Delta \ell\) that encodes the geometric and photometric differences between the light sources (e.g., direction, intensity, temperature). 
Our relighting model, \textbf{LightCtrl}, then synthesizes the appearance of the source object under the target illumination:
\begin{equation}
\hat{x}_s^{\,\ell_t} \;=\; f_\theta\!\big(x_s^{\,\ell_s},\, \Delta\ell\big).
\label{eq:method-overview}
\end{equation}
This formulation explicitly conditions the model on the source appearance and the reference-derived lighting shift, enabling controllable and consistent relighting.

Prior approaches often estimate full intrinsic components as diffusion conditioning, requiring dense G-buffer supervision and complex inverse-rendering optimization pipelines that frequently miss high-frequency effects. In addition, these methods typically rely on curated assets with physically modeled materials, making data acquisition expensive and limiting scalability.

Sec.~\ref{sec:fewshot} introduces a few-shot latent proxy that provides compact material–geometry priors from sparse PBR signals.  
Sec.~\ref{sec:mask} presents a lighting-aware mask that identifies sensitive lighting regions and guides spatial conditioning during denoising.  
Sec.~\ref{sec:control} describes a DPO-based post-training stage that refines the proxy branch for improved physical consistency under sparse supervision.  
Finally, Sec.~\ref{sec:dataset} details the construction of our illumination-controlled object-level dataset, which enables scalable, diverse, and physically consistent training.

\subsection{Few-shot Latent Proxy Conditioning}\label{sec:fewshot}

Single-image relighting is severely under constrained, and diffusion models often alter geometry or material appearance when illumination changes are large. Intrinsic-based two-stage pipelines alleviate this issue but require dense G-buffer supervision and suffer from stage misalignment, while purely latent approaches provide little physical structure and thus weak controllability. We seek an alternative that retains physical cues without full intrinsic reconstruction.

To this end, we introduce a lightweight encoder–decoder $E_\phi$ that predicts a compact \emph{latent proxy} 
\(\hat{\mathcal{B}}=\{a,n,r,m\}\in\mathbb{R}^{H\times W\times 8}\)
containing albedo, normals, roughness, and metallicity from the source image \(x_s^{\ell_s}\).  
Unlike full intrinsic recovery, the proxy is learned in a \emph{few-shot} manner: only a small subset of training images contains PBR supervision, and the proxy branch is updated using the loss
\begin{equation}
\label{eq:proxy-loss}
\mathcal{L}_{\text{proxy}} =
\begin{aligned}[t]
&\lambda_a \|a-\hat{a}\|_1
+ \lambda_n \bigl(1-\langle n,\hat{n}\rangle\bigr) \\
&\quad
+ \lambda_r \|r-\hat{r}\|_1
+ \lambda_m\,\mathrm{BCE}(m,\hat{m}),
\end{aligned}
\end{equation}
which captures the natural smoothness of albedo/roughness, unit-normal consistency, and quasi-binary metallic behavior.  
This sparse supervision stabilizes the proxy while allowing the main diffusion model to train on the full unlabeled dataset.

The predicted proxy maps are then spatially pooled and projected into a single conditioning token
\begin{equation}
t_{\text{proxy}} = f_{\text{proj}}(E_\phi(x_s^{\ell_s})) \in \mathbb{R}^{1\times768},
\end{equation}
which is injected into the denoiser alongside appearance and lighting tokens.  
This latent proxy token supplies material and geometry-aware priors that constrain the denoising trajectory, providing the physical structure needed for precise illumination control without requiring full intrinsic decomposition.

\subsection{Lighting-Aware Mask Prediction}\label{sec:mask}

Illumination changes typically affect only a small subset of pixels, such as shadow boundaries or specular regions, while most areas preserve the intrinsic appearance. Without spatial guidance, a diffusion model tends to distribute edits across the entire image, potentially altering albedo-consistent regions and destabilizing geometry. To address this, we introduce a \emph{lighting-aware mask} that highlights regions where illumination-driven changes are expected.

Given a source--target pair $(x_s^{\ell_s}, x_r^{\ell_t})$, we derive a soft ground-truth mask $M_{\mathrm{gt}}$ based on radiometric differences in linear luminance $Y_s$ and $Y_t$:
\begin{equation}
M_{\mathrm{gt}} = 
\mathcal{N}\!\left(
\alpha\, \bigl|\log Y_t - \log Y_s\bigr| 
+ (1-\alpha)\, D_{\mathrm{robust}}(Y_s, Y_t)
\right),
\end{equation}
where $D_{\mathrm{robust}}$ compensates for exposure variations and $\mathcal{N}(\cdot)$ normalizes the result into a stable $[0,1]$ soft mask. This yields a concise estimate of illumination-sensitive regions while suppressing texture-only differences.

At training time, the model cannot access the target image, so a lightweight predictor $m_\theta$ infers the mask from the source appearance and the relative lighting encoding:
\begin{equation}
M_\theta = m_\theta(x_s^{\ell_s}, \Delta\ell).
\end{equation}
The predictor is trained using a combination of binary cross-entropy and Dice loss against $M_{\mathrm{gt}}$.

To emphasize illumination-variant regions during diffusion, we transform the ground truth mask into a spatial weight map $W$, which modulates the noise reconstruction loss. This encourages the denoiser to allocate greater capacity to pixels influenced by lighting changes while preserving stability in illumination-invariant regions.

\subsection{Post-Training for Latent Encoder}\label{sec:control}

While the few-shot proxy provides essential material–geometry cues, the encoder is still trained from sparse PBR supervision and may lack the reliability required for precise illumination control. In particular, errors in albedo or normals can propagate into the conditioning pathway and lead to inconsistent relighting. To stabilize the proxy, we perform a \emph{DPO-style post-training} stage that refines the PBREncoder $E_\phi$ while freezing the main backbone.

For each supervised sample, the GrounTruth PBR maps $y_{\text{pos}}$ serve as the preferred target, and the current encoder output $y_{\text{neg}} = E_\phi(x_s^{\ell_s})$ provides a less-preferred alternative. We compute a physics-based reward difference
\begin{equation}
\Delta r = r(y_{\text{pos}}) - r(y_{\text{neg}}),
\end{equation}
where $r(\cdot)$ aggregates albedo and roughness $L_1$, normal angular deviation, and metallic BCE. 

A frozen reference encoder $E_{\phi^{\mathrm{ref}}}$ provides stable likelihood estimates for $(y_{\text{pos}}, y_{\text{neg}})$, and the PBREncoder is updated using a DPO objective that increases the likelihood of higher-reward predictions. This post-training step strengthens the physical consistency of the latent proxy under sparse supervision, improving the stability and controllability of downstream relighting without modifying the diffusion model itself.

Given the noisy latent $z_t$ at timestep $t$, the denoiser is conditioned on 
the source appearance token $t_{\mathrm{img}}$, the relative-lighting token $t_{\mathrm{light}}$, 
and the few-shot proxy token $t_{\mathrm{phys}}$, which respectively encode appearance, target illumination, 
and lightweight material–geometry cues.  
Our final diffusion objective is
\begin{equation}
\label{eq:final-diffusion-loss}
\mathcal{L}_{\mathrm{diff}}
=
\big\|
W \odot 
\bigl(
\epsilon -
\epsilon_\theta(
z_t,\, t \mid
t_{\mathrm{img}},\, t_{\mathrm{light}},\, t_{\mathrm{phys}}
)
\bigr)
\big\|_2^2 ,
\end{equation}
where $\epsilon$ is the ground truth noise and $W$ is the lighting-aware spatial weight map 
that emphasizes illumination-sensitive regions predicted by our mask module.

\section{Dataset}\label{sec:dataset}

To enable controllable relighting and physically grounded appearance learning, we introduce \textbf{ScaLight}, a large-scale synthetic dataset of objects rendered under systematically varied illumination. Unlike existing scene-level datasets with fixed or weakly controlled lighting, ScaLight focuses on per-object rendering with consistent geometry and material across lighting changes, providing clean supervision for learning reflectance–shading behavior. Objects are procedurally sampled from diverse 3D asset repositories~\cite{objaverseXL, deitke2022objaverseuniverseannotated3d, collins2022abodatasetbenchmarksrealworld, khanna2023habitatsyntheticscenesdataset} and rendered using a physically based pipeline with configurable directional, point, and environment lights. This fully automated setup scales easily and enables efficient generation of high-quality relighting pairs with minimal human intervention.

\begin{table}[t]
  \centering
  \normalsize
  \caption{\textbf{Comparison with other object-level datasets.} \textbf{ScaLight} features large-scale scalability, detailed \textbf{camera and light metadata} (recording exact camera and illumination parameters), and fully controllable \textbf{lighting setups} (supporting diverse and flexible lighting effects). Symbols: \cmark~full; \lmark~limited; \xmark~none.}
  \label{tab:obj_level_compare}
  \vspace{0.25em}
  \resizebox{1.0\linewidth}{!}{
  \begin{tabular}{@{\centering}lp{2cm}<{\centering}p{1.5cm}<{\centering}p{1.5cm}<{\centering}p{1.5cm}<{\centering}@{}}
    \toprule
    \textbf{Dataset} & \textbf{\#Objects} & \textbf{Lighting Control} & \textbf{Pose / Light} \\
    \midrule
    OpenIllumi~\cite{liu2023openillumination}        & 64    & \lmark\  & \xmark \\
    OWL~\cite{ummenhofer2024objectslightingrealworlddataset}   & 8     & \lmark\  & \lmark\ \\
    Rene~\cite{Toschi_2023_CVPR}               & 20  & \lmark          & \xmark  \\
    LightProp~\cite{he2024neurallightrigunlockingaccurate}               & 80K+  & \cmark                 & \xmark         \\
    \midrule
    \textbf{ScaLight (Ours)}& \textbf{300K+} & \textbf{\cmark} & \textbf{\cmark\ } \\
    \bottomrule
  \end{tabular}
  }
  \vspace{-1.5em}
\end{table}

We randomly sample multiple camera viewpoints on a hemisphere around each object and render every view under a variety of lighting configurations by perturbing the position, orientation, energy, temperature, and color of light sources. Each rendered frame is accompanied by complete metadata, including camera pose and all illumination parameters, which supports both relighting tasks and explicit illumination disentanglement. Figure~\ref{fig:dataoverview} provides an overview of this rendering process, where controlled camera–light sampling yields consistent, multi-view, multi-illumination observations. Formally, each rendered sample in ScaLight can be represented as
\begin{equation}
\mathcal{D} = \{ (x_i^{\ell_j},\, \pi_i,\, \ell_j) \mid i = 1{\ldots}N_o,\, j = 1{\ldots}N_\ell \}, 
\label{eq:data}
\end{equation}

\begin{table}[t]
  \centering
  \caption{Scene-level relighting comparison on the \textbf{MIIW} testset. 
  \textbf{Bold} indicates the best performance.}
  \label{tab:scene_miiw}
  \normalsize 
  \setlength{\tabcolsep}{12pt} 
  
  \begin{tabular}{@{}lccc@{}}
    \toprule
    \textbf{Method} & \textbf{RMSE} $\downarrow$ & \textbf{SSIM} $\uparrow$ & \textbf{PSNR} $\uparrow$ \\
    \midrule
    RGB$\leftrightarrow$X~\cite{zeng2024rgb}  & 0.389 & 0.425 & 8.25 \\
    IC-Light~\cite{zhang2025scaling}          & 0.413 & 0.337 & 7.94 \\
    LumiNet~\cite{xing2025luminetlatentintrinsicsmeets} & \textbf{0.139} & \textbf{0.904} & 17.20 \\
    \midrule
    \textbf{Ours}                             & 0.167 & 0.655 & \textbf{18.30} \\
    \bottomrule
  \end{tabular}
  \vspace{-1.5em}
\end{table}

\begin{table*}[t]
\centering
\caption{\textbf{Quantitative comparison on relit color output under three lighting variations.}
All metrics are computed on the normalized RGB range $[-1,1]$. 
\emph{Temperature}: correlated color temperature (CCT) shift; 
\emph{Position}: light direction/placement change (yaw/pitch); 
\emph{Energy}: light intensity scaling.}
\vspace{0.4em}
\begingroup
\setlength{\tabcolsep}{5pt} 
\renewcommand{\arraystretch}{1.05}
\resizebox{\linewidth}{!}{%
\begin{tabular}{l|c|ccc|ccc|ccc}
\toprule
\multirow{2}{*}{\textbf{Method}} & \multirow{2}{*}{\textbf{Label}} 
& \multicolumn{3}{c|}{\textbf{Temperature} (CCT shift)} 
& \multicolumn{3}{c|}{\textbf{Position} (yaw/pitch)} 
& \multicolumn{3}{c}{\textbf{Energy} (intensity)} \\
\cmidrule(lr){3-5}\cmidrule(lr){6-8}\cmidrule(lr){9-11}
& & \textbf{RMSE$\downarrow$} & \textbf{SSIM$\uparrow$} & \textbf{PSNR$\uparrow$}
  & \textbf{RMSE$\downarrow$} & \textbf{SSIM$\uparrow$} & \textbf{PSNR$\uparrow$}
  & \textbf{RMSE$\downarrow$} & \textbf{SSIM$\uparrow$} & \textbf{PSNR$\uparrow$} \\
\midrule
IC-Light~\cite{zhang2025scaling}   & text   & 0.397 & 0.346 & 8.21  & 0.375 & 0.363 & 8.65  & 0.380 & 0.382 & 8.63 \\
DreamLight~\cite{dreamlight} & text   & 0.349 & 0.559 & 9.40  & 0.221 & 0.662 & 13.3  & 0.221 & 0.671 & 13.5 \\
x2rgb~\cite{zeng2024rgb}        & G-buffer & 0.472 & 0.493 & 7.81  & 0.487 & 0.592 & 8.32  & 0.484 & 0.676 & 8.89 \\
LBM~\cite{chadebec2025lbm}        & image  & 0.064 & 0.957 & 27.8  & 0.084 & 0.936 & 23.1 & 0.073 & 0.956 & 25.3 \\
Luminet~\cite{xing2025luminetlatentintrinsicsmeets}        & image  & 0.172 & 0.848 & 15.8  & 0.146 & 0.847 & 17.8  & 0.164 & 0.825 & 16.2 \\
\midrule
Ours (w/o proxy)   & Light Info & 0.062 & 0.953 & 28.1  & 0.087 & 0.931 & 22.4  & 0.171 & 0.902 & 18.0 \\
Ours (w/o mask)    & Light Info & 0.073 & 0.937 & 27.8 & 0.102 & 0.865 & 20.5  & 0.126 & 0.907 & 23.2 \\
Ours (w/o dpo)        & Light Info & 0.114 & 0.840 & 22.8 & 0.163 & 0.854 & 19.8  & 0.194 & 0.825 & 17.5 \\
Ours (full)        & Light Info & \textbf{0.053} & \textbf{0.974} & \textbf{30.2} 
                                   & \textbf{0.074} & \textbf{0.929} & \textbf{25.6}
                                   & \textbf{0.083} & \textbf{0.988} & \textbf{27.1} \\
\bottomrule
\end{tabular}%
}
\endgroup
\label{tab:relighting_variations}
\end{table*}

where $x_i^{\ell_j}$ denotes the rendering of object $i$ under illumination $\ell_j$, and $\pi_i$ and $\ell_j$ represent the associated camera pose and light configuration.  
A small subset of objects additionally includes material annotations $(a_i, n_i, r_i, m_i)$ for weak supervision, enabling few-shot learning of intrinsic cues in our proxy encoder.  
This formulation naturally supports sampling of relighting pairs $(x_i^{\ell_s}, x_i^{\ell_t})$ with a known illumination difference $\Delta\ell=\ell_t-\ell_s$, which we use for both supervised and self-supervised objectives.

ScaLight contains over \textbf{300K} controllable 3D objects and more than \textbf{1M} rendered images (details in the appendix), covering a wide range of lighting variations.  
For each object, we sample multiple viewpoints and render it under diverse combinations of directional, point, and area lights~\cite{blender}, varying light position, orientation, intensity, color, and temperature.  
This systematic parameterization provides dense illumination sampling while preserving fixed geometry and materials, making ScaLight a scalable and physically consistent benchmark for relighting.  
Table~\ref{tab:obj_level_compare} further compares ScaLight with existing object-level datasets, highlighting its substantially larger scale and richer illumination diversity. To evaluate generalization beyond controlled synthetic settings, we additionally incorporate the MIIW scene-level relighting dataset~\cite{murmann19}.  Together, ScaLight and MIIW allow LightCtrl to be trained and evaluated under both controlled multi-illumination supervision and realistic real-world conditions.

\section{Experiment}
\begin{figure}[t]
    \centering
    \includegraphics[width=\linewidth]{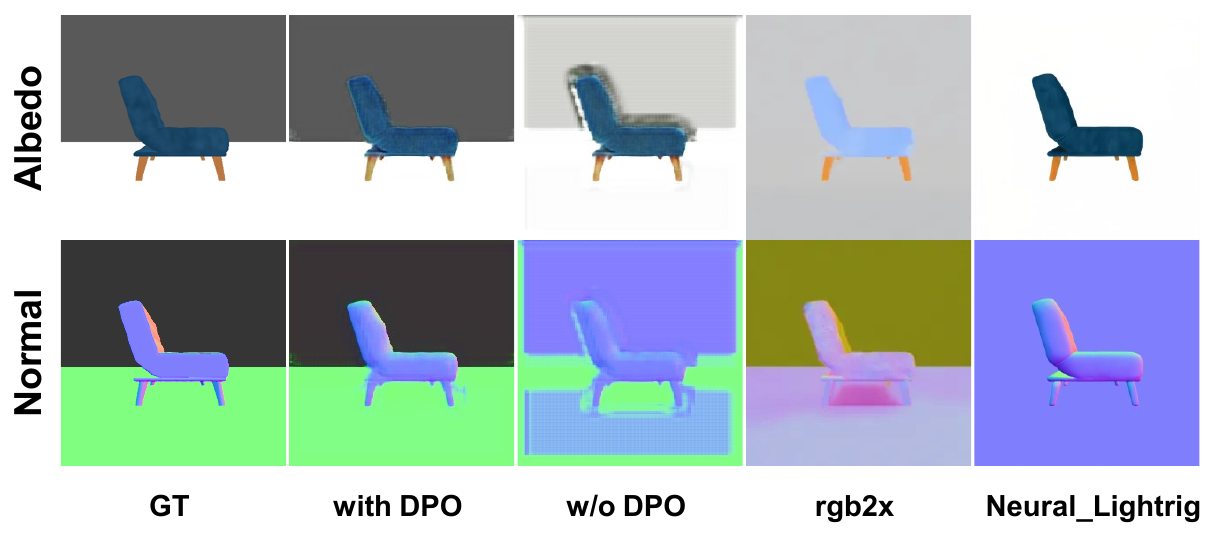}
    \caption{
    Visual comparison of intrinsic decomposition on a chair object.
    From left to right: ground-truth (GT), our method with DPO, our method without DPO, rgb2x~\cite{zeng2024rgb}, and Neural\_Lightrig~\cite{he2024neurallightrigunlockingaccurate}.
    The DPO fine-tuning clearly suppresses artifacts and produces more accurate albedo and normal predictions, demonstrating the effectiveness of our DPO strategy.
    }
    \label{fig:dpo_ablation_chair}
    \vspace{-1em}
\end{figure}

\begin{figure}[t]
    \centering
    \includegraphics[width=\linewidth]{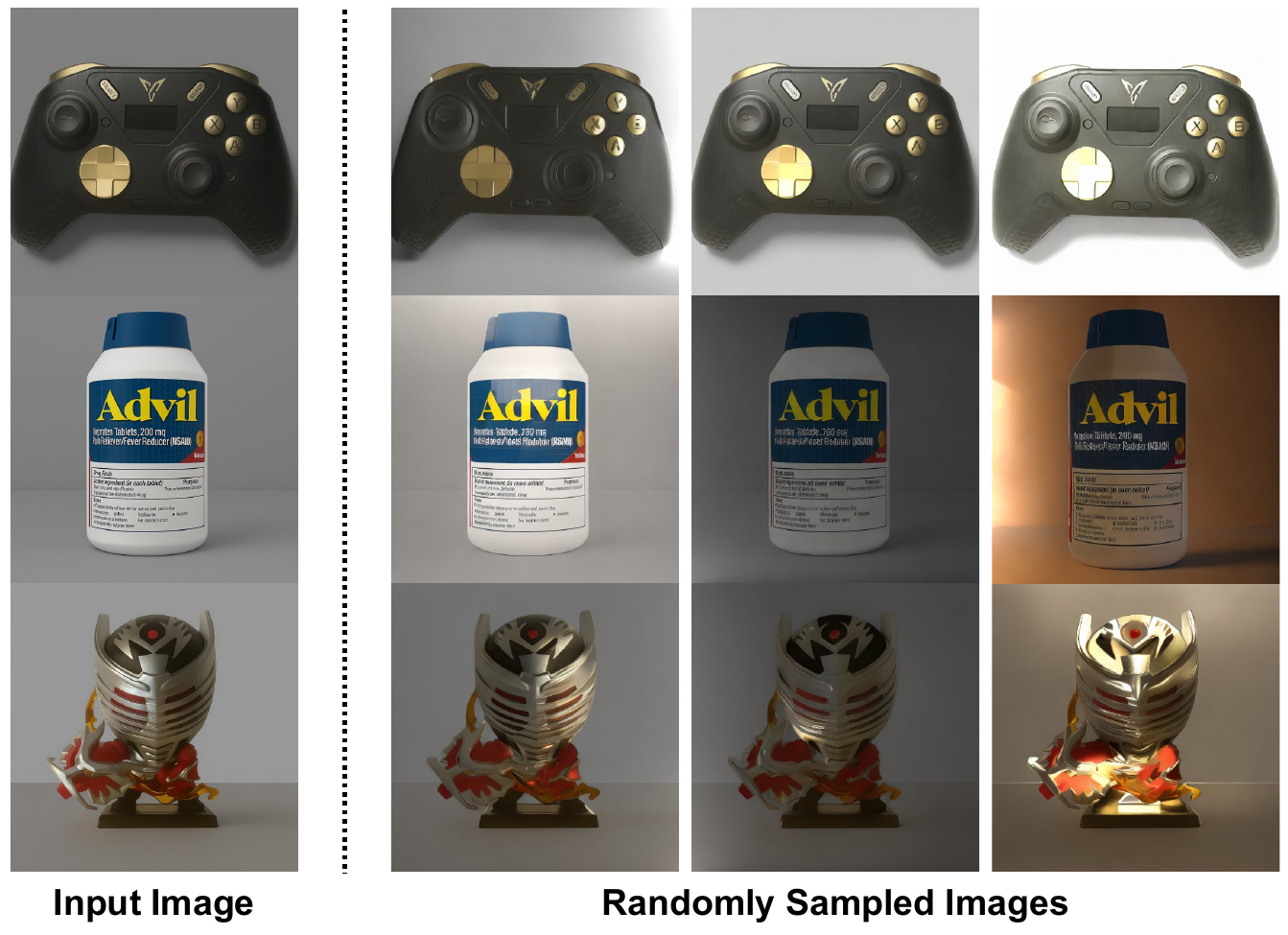}
    \caption{
    Qualitative results on in-the-wild images. 
    Our model produces realistic relighting on Internet and product photos (right). Hand-captured examples confirm robust generalization to uncontrolled lighting.
    }
    \label{fig:in_the_wild_random}
    \vspace{-1.5em}
\end{figure}

\begin{figure*}[t]
    \centering
    \includegraphics[width=\textwidth]{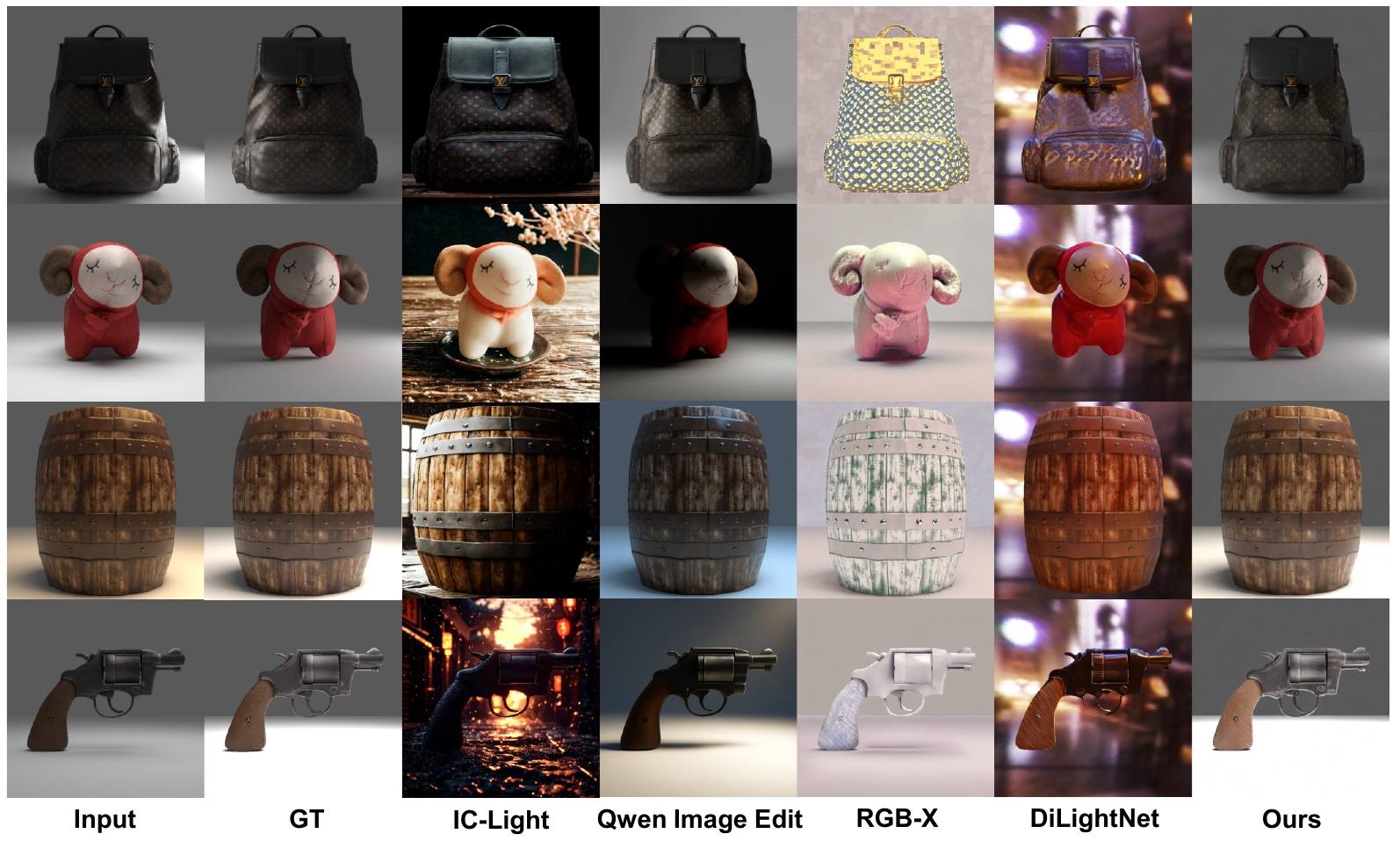}
    \vspace{-0.2in}
    \caption{
    Object-level relighting comparison. 
    Given a single input image (left), we compare our method against IC-Light, Qwen Image Edit, RGB-X, and DiLightNet. 
    Competing methods often introduce strong color shifts, texture distortions, or inconsistent shading, while our approach produces well-relighting results that preserve materials, geometry, and object identity. 
    These examples demonstrate our model’s superior ability to perform physically plausible object-centric relighting.
    }
    \label{fig:object_level_comparison}
        \vspace{-0.1in}
\end{figure*}

We evaluate LightCtrl on both synthetic and real-world datasets to assess its performance in controllable relighting and illumination transfer.  In \textbf{Sec.~\ref{sec:quan}}, we report standard quantitative metrics (RMSE, SSIM, PSNR) under diverse lighting changes to measure photometric accuracy.  
\textbf{Sec.~\ref{sec:qual}} presents qualitative comparisons on synthetic and in-the-wild images, highlighting visual fidelity and fine-grained illumination control.  \textbf{Sec.~\ref{sec:ablation}} provides ablation studies examining the contribution of each component in our framework, including the few-shot latent proxy and lighting-aware conditioning.

\subsection{Quantitative Evaluation}\label{sec:quan}
Since existing benchmarks lack precise illumination control, we evaluate on a controlled subset of \textbf{ScaLight}, which provides physically consistent lighting and material annotations. We compute metrics on a held-out set of 1.5K unseen objects rendered under diverse lighting directions, intensities, and color temperatures. 
Following standard practice, we report PSNR, RMSE, and SSIM to assess relighting accuracy. As shown in Table~\ref{tab:relighting_variations}, our method achieves superior overall quantitative performance. Notably, while LumiNet attains a higher SSIM score, this discrepancy reflects a fundamental perception-distortion trade-off. Pixel-aligned metrics like SSIM inherently favor conservative structural preservation, often rewarding models that retain original shading or produce blurry averages (akin to style transfer). In contrast, LightCtrl executes flexible and precise illumination commands that require significant structural changes, such as synthesizing novel, sharp cast shadows. While this physically correct behavior naturally incurs an SSIM penalty\cite{Blau_2018_CVPR} compared to conservative approaches, our method ultimately achieves more photorealistic and photometrically consistent relighting. Overall, these results demonstrate the effectiveness of our approach while requiring only sparse PBR supervision during training.

Beyond object-level evaluation, we further validate scene-level performance on 
indoor benchmarks with complex geometry and cluttered layouts (e.g., MIIW~\cite{murmann19}). 
These datasets contain real photographs captured under diverse illumination 
setups, providing a challenging testbed for global light transport, shadows, 
and interreflections. We report scene-level RMSE, SSIM, and PSNR across all 
methods, and show that our model generalizes from object-centric training to 
full scenes, consistently outperforming intrinsic-based and diffusion baselines 
in both relighting accuracy and perceptual quality.

\begin{table}[t]
  \centering
  \vspace{-0.4em}
  \small
  \renewcommand{\arraystretch}{1.1}
  \setlength{\tabcolsep}{6pt}
    \caption{\textbf{User preference study ($N=35$).} We evaluate human preference separately on complex real-world scenes (MIIW and RWR) and controlled objects (ScaLight). Our method significantly outperforms existing baselines, achieving the highest preference rate in both settings.}
  \begin{tabular*}{\linewidth}{@{\extracolsep{\fill}}c|cc@{}}
    \hline
    \multirow{2}{*}{\textbf{Method}} & \multicolumn{2}{c}{\textbf{User Preference Rate}~($\uparrow$)} \\
    & \textbf{Scene-Level} & \textbf{Object-Level} \\
    \hline
    LumiNet~\cite{xing2025luminetlatentintrinsicsmeets}  & 12.83\% & 4.3\% \\
    IC-Light~\cite{zhang2025scaling} & 9.37\% & 11.45\% \\
    NanoBanana & 22.07\% & 2.8\% \\
    \hline
    \textbf{Ours} & \textbf{55.73\%} & \textbf{81.45\%} \\
    \hline
  \end{tabular*}
  \vspace{-0.5em}
  \label{tab:user_study}
  \vspace{-1.0em}
\end{table}

\begin{figure*}[t]
    \centering
    \includegraphics[width=\textwidth]{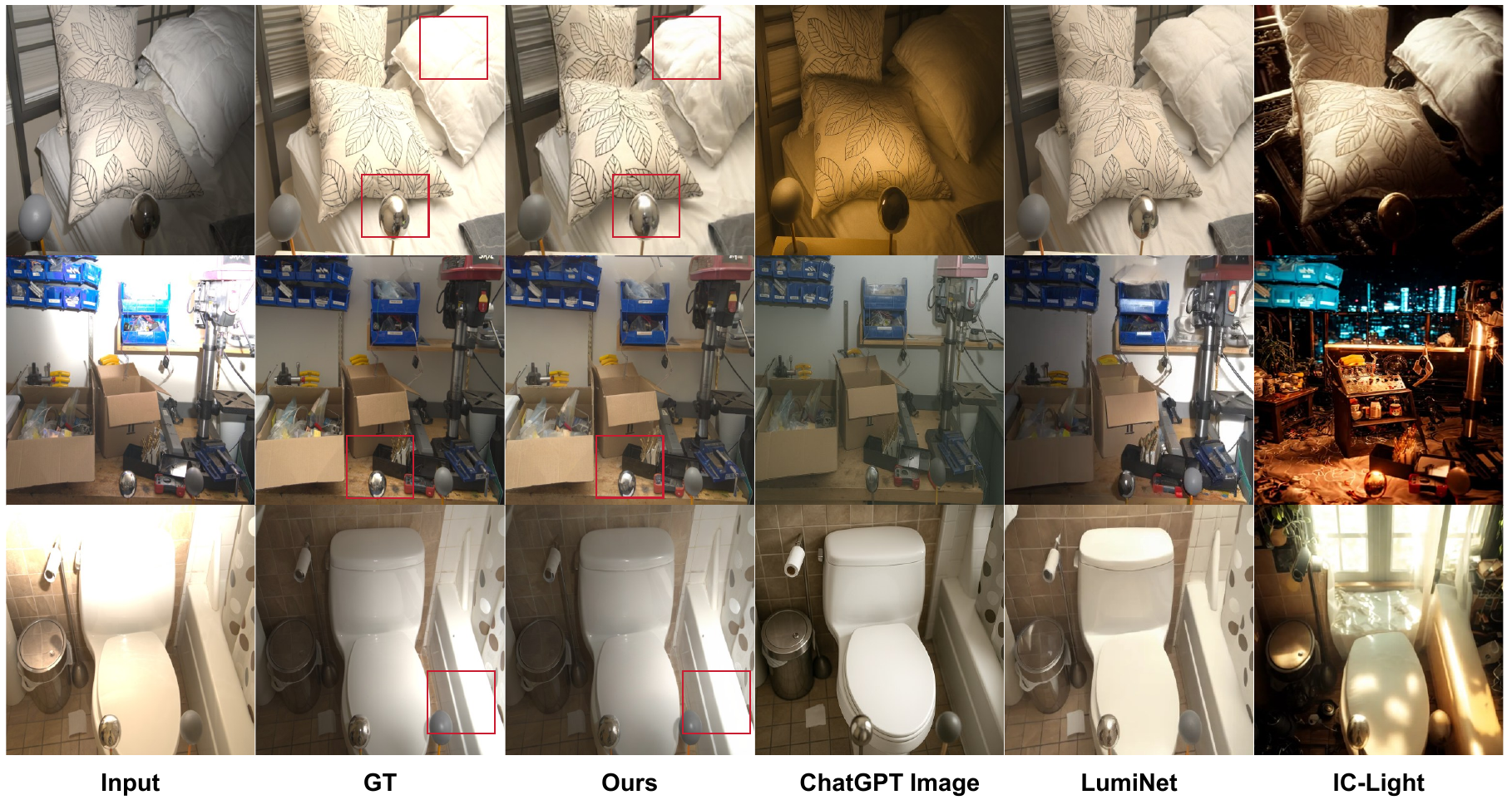}
           \vspace{-0.1in}
    \caption{
    Scene-level relighting comparison on indoor environments. 
    From left to right: input, ground-truth relighting, our method, ChatGPT Image, and LumiNet.
    Our approach produces realistic global illumination and consistent local shading, especially around the reflective light probes highlighted by red boxes, 
    whereas the generative baselines tend to make color shifts or over-smoothed shading.
    }
    \label{fig:scene_level_comparison}
               \vspace{-0.1in}
\end{figure*}

\subsection{Qualitative Evaluation}
\label{sec:qual}

We first evaluate LightCtrl on a held-out subset of \textbf{ScaLight}, which provides object-centric images rendered under systematically varied illumination.  
Across changes in light direction, intensity, and color temperature, LightCtrl preserves shading discontinuities, specular structure, and fine-scale material cues, while maintaining consistent albedo appearance.  
Compared with intrinsic-based approaches~\cite{zeng2024rgb, jin2024neuralgafferrelightingobject} and latent-only diffusion models~\cite{zhang2025scaling, xing2025luminetlatentintrinsicsmeets}, we observe fewer color shifts, reduced oversmoothing, and more stable reproduction of high-frequency surface details.  
These results suggest that the combination of the few-shot latent proxy and mask-guided conditioning provides effective spatial guidance for modeling illumination changes at fine granularity.

To assess generalization beyond object-centric settings, we further evaluate on the \textbf{MIIW} dataset, which contains real indoor scenes with complex geometry, clutter, and mixed materials.  
Despite being trained primarily on synthetic object-level data, LightCtrl produces illumination adjustments that are consistent with global scene structure, with shadows and highlights adapting coherently across surfaces.  
In comparison, existing relighting methods such as IC-Light~\cite{zhang2025scaling}, RGB2X~\cite{zeng2024rgb}, and LumiNet~\cite{xing2025luminetlatentintrinsicsmeets} exhibit more frequent artifacts—including shadow discontinuities or attenuation of surface texture—under the same evaluation protocol.  
These observations indicate that it transfers illumination behavior to novel real scenes more reliably.

Finally, we test LightCtrl on \textbf{in-the-wild images} captured with handheld mobile devices, which exhibit uncontrolled backgrounds, clutter, and varying exposure.  
Across diverse examples, our model generates relit outputs with plausible shadow relocation and highlight behavior, while maintaining overall material appearance.  
Although these settings are substantially farther from the training distribution, LightCtrl retains stable performance, demonstrating robustness under real-world variability. To quantitatively validate this perceptual realism, we conducted a comprehensive user preference study. As detailed in Table \ref{tab:user_study}, participants overwhelmingly favored our approach over state-of-the-art baselines, with LightCtrl achieving the highest preference rates in both controlled object level (81.45\%) and complex scene level (55.73\%) evaluations.

\subsection{Ablation Study}\label{sec:ablation}
We analyze the impact of each component in LightCtrl, namely the few-shot latent proxy, the lighting-aware mask, and the DPO post-training stage. Object-level results under different illumination variations are summarized in Table~\ref{tab:relighting_variations}, and intrinsic predictions are visualized in Fig.~\ref{fig:dpo_ablation_chair}.

\textbf{Few-shot Latent Proxy.}
Removing the proxy (\emph{w/o proxy}) leads to consistent degradation in RMSE and PSNR across temperature, position, and energy changes, with the largest gaps appearing under color-temperature and intensity shifts. SSIM also decreases for temperature and energy, while remaining comparable for position. These trends indicate that the proxy supplies material and geometry-dependent cues that help the model maintain stable shading and reflectance when illumination varies.

\textbf{Lighting-aware Mask.}
Without the mask (\emph{w/o mask}), performance drops on all three lighting variations: RMSE increases and both SSIM and PSNR decrease relative to the full model. Visually, the model tends to over-edit regions whose appearance should remain invariant, leading to smoother albedo and weakened shadow boundaries. The mask thus provides effective spatial guidance, steering the denoiser toward illumination-sensitive pixels while better preserving appearance-invariant areas.

\textbf{DPO Post-training.}
Disabling DPO (\emph{w/o dpo}) results in the largest performance loss among all ablations, with noticeably higher RMSE and lower PSNR/SSIM for every lighting variation. As shown in Fig.~\ref{fig:dpo_ablation_chair}, the encoder without DPO produces noisier albedo and distorted normals, whereas DPO-refined proxies are cleaner and more physically plausible. This confirms that the post-training stage is crucial for stabilizing intrinsic cues under sparse supervision and improving downstream relighting quality.

Overall, these ablations confirm that each component addresses a distinct challenge, proxy guidance for material awareness, masking for spatial selectivity, and DPO for intrinsic stability, and that all three contribute synergistically to LightCtrl’s relighting performance.

\section{Limitations}

Although LightCtrl improves controllable relighting under diverse conditions, its strongest evidence is still concentrated in synthetic, object-centric settings. When transferred to cluttered real scenes with complex global illumination, the control behavior becomes less robust and the quality gain is more mixed. In particular, performance on real-scene benchmarks reflects a practical trade-off: the model preserves stable appearance cues in many cases, but it is still sensitive to long-range light transport and strong specular interactions.

\begin{figure}[t]
    \centering
    \includegraphics[width=0.9\linewidth]{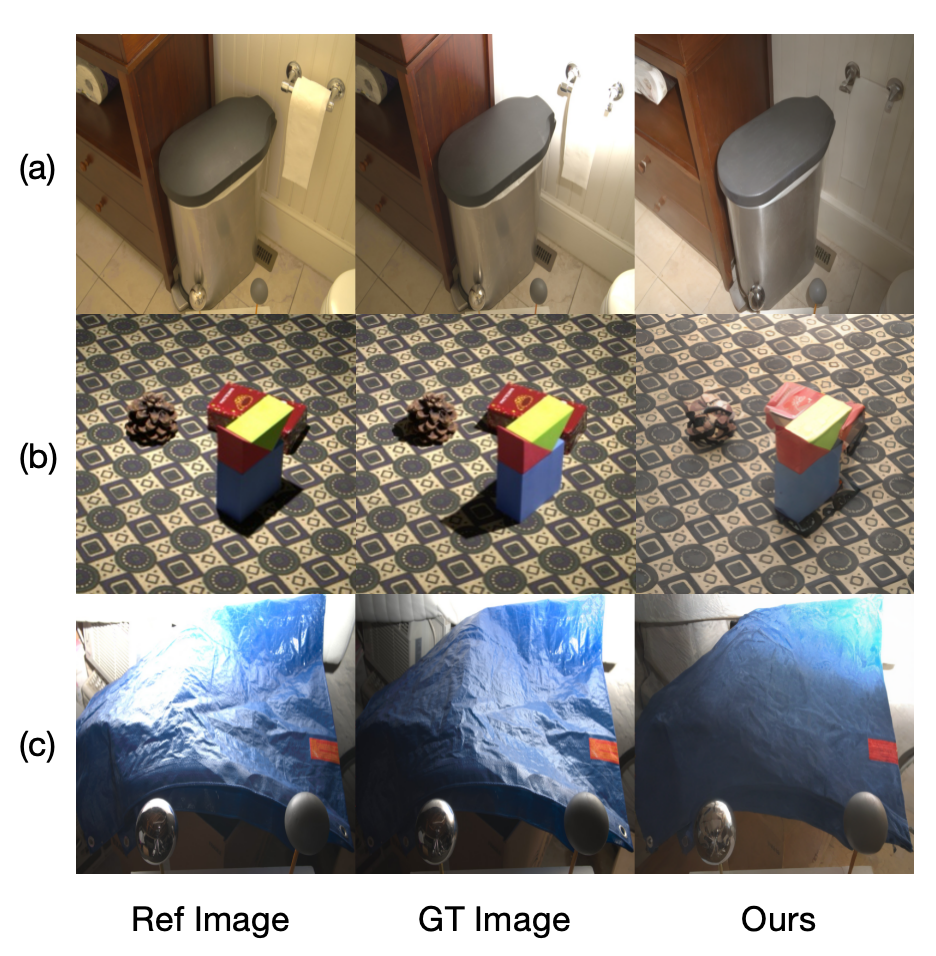}
    \caption{
    Targeted real-scene failure analysis on MIIW and RWR~\cite{yang2025relighting}, showing representative errors in shadow recasting, structure preservation, and highlight-heavy regions.
    }
    \label{fig:failure_cases}
    \vspace{-1.5em}
\end{figure}

As shown in Fig.~\ref{fig:failure_cases}, a primary failure mode is inaccurate global cast-shadow recasting. The model can usually handle local shading changes, but it may miss sharp long-range shadows between distant objects and supporting surfaces, especially under large viewpoint-dependent lighting changes. This is consistent with our proxy design: because the conditioning signal is inferred from sparse single-view cues instead of dense multi-view geometry or explicit occlusion reasoning, the denoiser may default to smooth ambient adjustments rather than physically precise shadow intersections.

A second limitation appears in high-frequency geometry and highlight-heavy regions. Under strong illumination contrast or concentrated specular highlights, fine structural details can be over-smoothed, and local textures may be partially washed out. In these cases, the latent proxy does not always provide enough high-frequency constraints to separate true geometry variation from illumination variation, which leads to flattened relief and detail loss around bright reflections. Despite these issues, color consistency is comparatively stable, suggesting that the proxy still anchors albedo better than purely latent editing baselines.

Future work will focus on improving robustness in exactly these regimes by strengthening global light-transport reasoning, introducing richer geometry/occlusion supervision, and adding control-specific diagnostics for shadow accuracy, structural fidelity, and highlight preservation. We believe this direction is important for narrowing the remaining gap between object-centric training and reliable real-scene controllability.

{
    \small
    \bibliographystyle{ieeenat_fullname}
    \bibliography{main}
}

\clearpage

\appendix


This section provides additional implementation details for LightCtrl that are
omitted from the main paper for clarity. We first describe how the relative
illumination encoding $\Delta\ell$ is constructed from physically meaningful
lighting parameters, with emphasis on the conversion from user-specified
yaw--pitch edits to the spherical-harmonic (SH) representation used by the
model. We then elaborate on the design of the lighting-aware mask and its
conversion into a spatial weighting scheme for denoising. Finally, we detail
how appearance, lighting, and proxy tokens are integrated into the conditioning
architecture to achieve controllable relighting.

\section{Relative Illumination Encoding $\Delta\ell$}
\label{subsec:delta_l}

Our illumination encoding is designed to support intuitive user control while
remaining stable for learning. In practice, lighting edits are specified in a
2D \emph{yaw--pitch} space---a natural interface for rotating a directional
light around an object. However, diffusion models operate on vector-valued
embeddings, not angles. We therefore convert yaw--pitch edits into a
physically meaningful representation through the following steps:

\paragraph{Step 1: From yaw--pitch control to 3D light direction.}
Given a source lighting configuration with angular parameters
$(\theta_s, \phi_s)$ and a user-specified edit defining a target illumination
$(\theta_t, \phi_t)$, we map each pair into a unit direction vector in the
camera coordinate frame:
\[
\boldsymbol{\omega}(\theta,\phi)
=
\begin{bmatrix}
\cos\theta \,\sin\phi \\
\sin\theta \,\sin\phi \\
\cos\phi
\end{bmatrix}.
\]
This produces the source and target directions
$\boldsymbol{\omega}_s$ and $\boldsymbol{\omega}_t$, whose difference encodes
how illumination rotates around the object.

\paragraph{Step 2: SH projection of lighting direction.}
To obtain a smooth, rotation-aware embedding, we project each direction onto a
real spherical-harmonic (SH) basis up to second order ($l\le2$), yielding
$9$ coefficients per light:
\[
\mathbf{s}_{\mathrm{SH}}(\boldsymbol{\omega})
=
\bigl[
Y_0^0(\boldsymbol{\omega}),
Y_1^{-1}(\boldsymbol{\omega}),\dots,
Y_2^{2}(\boldsymbol{\omega})
\bigr].
\]

Low-order SH provides a continuous angular representation in which small
yaw--pitch edits correspond to small changes in coefficient space. This
avoids the discontinuities and periodic ambiguities associated with raw angle
differences.

\paragraph{Step 3: Constructing the relative lighting embedding.}
Since relighting depends only on \emph{how} illumination changes, not on its
absolute parameters, we use the difference between SH encodings:
\[
\Delta\mathbf{s}_{\mathrm{SH}}
=
\mathbf{s}_{\mathrm{SH}}(\boldsymbol{\omega}_t)
-
\mathbf{s}_{\mathrm{SH}}(\boldsymbol{\omega}_s).
\]
We compute photometric and chromatic differences in the same manner: the
relative intensity is simply the log-intensity difference
$\Delta \log E = \log E_t - \log E_s$, and the relative color temperature is
the normalized scalar difference $\Delta\tau = \tau_t - \tau_s$. These
quantities measure how much the lighting becomes brighter/dimmer or warmer/
cooler, without encoding their absolute values.
The full relative illumination vector is therefore
\[
\Delta\ell
=
\bigl[
  \Delta\mathbf{s}_{\mathrm{SH}},\;
  \Delta\log E,\;
  \Delta\tau
\bigr],
\]
which is mapped through a small MLP to produce the lighting token
$t_{\mathrm{light}}$.
This representation provides an intuitive parameterization for user edits
while remaining physically interpretable and numerically stable for diffusion
training.

\begin{figure}[t]
    \centering
    \includegraphics[width=\linewidth]{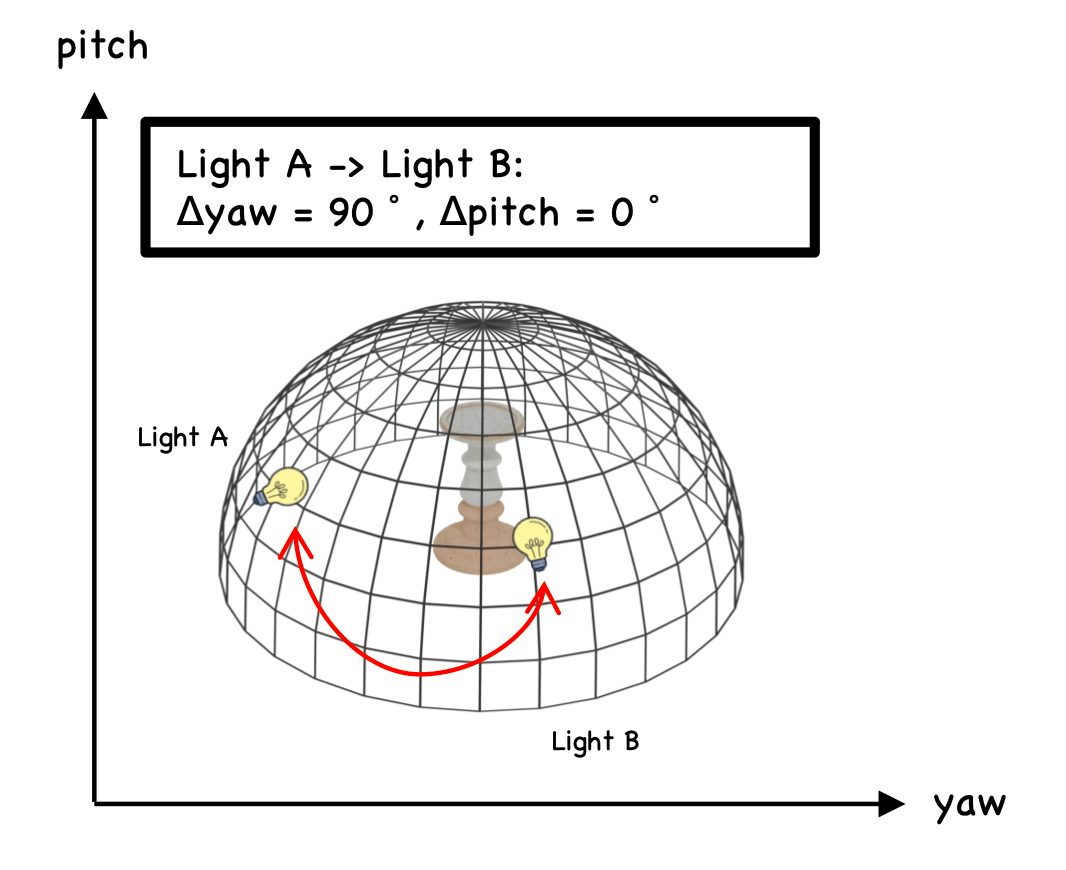}
    \caption{Illustration of our controllable lighting formulation.
    A lighting edit is first defined in yaw--pitch space (e.g.,
    Light~A $\rightarrow$ Light~B with $\Delta\mathrm{yaw}=90^\circ$,
    $\Delta\mathrm{pitch}=0^\circ$).
    These angular parameters are mapped to unit vectors on the hemisphere,
    then projected onto a low-order spherical-harmonic (SH) basis.
    The resulting SH difference $\Delta\mathbf{s}_{\mathrm{SH}}$ forms the
    directional component of our relative illumination encoding $\Delta\ell$.}
    \label{fig:light_control}
\end{figure}

\section{Lighting-aware Mask Design}
\label{subsec:mask_detail}
A key challenge in controllable relighting is that illumination changes are inherently spatially selective: specular highlights shift according to the light direction, shadows migrate across surfaces, and soft shading gradients adjust smoothly, while most of the object’s appearance—including albedo and fine geometry cues—remains unchanged. If a diffusion model is trained without spatial guidance, it tends to distribute lighting edits broadly across the image, unintentionally altering material properties or introducing inconsistent shading.

Changes in lighting should not prompt the model to modify intrinsic appearance. However, diffusion models often struggle to distinguish illumination-driven variations from texture changes, especially when trained on diverse lighting setups. A spatial mask offers a practical way to communicate “where the model should change things” versus “where it should stay invariant.” This encourages the model to treat material appearance as stable, while letting highlights, cast shadows, and shading gradients adapt fluidly to $\Delta\ell$.

Instead of relying on explicit geometric supervision or perfect shadow segmentation—which would require heavy manual annotation—our mask is derived directly from photometric observations. We compare the luminance fields of the source and target images under logarithmic scaling, which suppresses exposure differences and emphasizes shading variation. A robust, multi-kernel smoothing operation then extracts a soft spatial map that highlights areas where illumination behavior changes noticeably.

This produces a mask $M_{\mathrm{gt}}$ that is: smooth enough to guide diffusion without introducing high-frequency artifacts, responsive to both hard shadow boundaries and soft shading shifts, and fully generated without human annotation.

\section{Conditioning Architecture and Tokens}
\label{subsec:conditioning_tokens}

Our diffusion backbone follows a U-Net–based architecture with cross-attention
layers at each resolution. LightCtrl provides three complementary conditioning
signals---appearance, illumination, and proxy cues---which are injected into the
model through a unified token-based interface. Each token type is encoded into a
fixed-dimensional embedding and consumed by cross-attention during denoising.

\paragraph{Appearance token.}
The source image \(x_s^{\ell_s}\) is first encoded by a frozen
vision encoder \(F_{\mathrm{img}}\) (a CLIP-ViT variant), producing
\[
t_{\mathrm{img}} = F_{\mathrm{img}}(x_s^{\ell_s}) 
\;\in\; \mathbb{R}^{1 \times d},
\]
which captures the coarse geometry and view-dependent texture cues that should
remain invariant under relighting. This token serves as the primary identity
anchor throughout the diffusion process.

\paragraph{Lighting token.}
The relative illumination encoding \(\Delta\ell\)
(Sec.~\ref{subsec:delta_l}) is mapped through a lightweight MLP,
\[
t_{\mathrm{light}}
=
\mathrm{MLP}_{\ell}(\Delta\ell)
\;\in\; \mathbb{R}^{1 \times d},
\]
providing a compact representation of directional, energetic, and chromatic
lighting changes. This token determines the intended shading and highlight
appearance at each denoising timestep.

\paragraph{Proxy (PBR) token.}
The latent proxy estimated by the PBREncoder,
\(\hat{\mathcal{B}} = E_{\phi}(x_s^{\ell_s})\), is spatially pooled and
linearly projected:
\[
t_{\mathrm{pbr}}
=
W_{\mathrm{pbr}} \cdot 
\mathrm{Pool}\bigl(\hat{\mathcal{B}}\bigr)
\;\in\; \mathbb{R}^{1 \times d}.
\]
This token introduces soft material and geometry priors inferred from sparse
PBR supervision, compensating for the under-constrained nature of single-image
relighting.

\paragraph{Token fusion in cross-attention.}
At each U-Net block, the conditioning tokens are concatenated into a unified
sequence:
\[
T
=
\bigl[
t_{\mathrm{img}},
t_{\mathrm{light}},
t_{\mathrm{pbr}}
\bigr]
\;\in\; \mathbb{R}^{3 \times d}.
\]
Given a latent feature map \(z_t\) at diffusion step \(t\), the cross-attention
operation is:
\[
\mathrm{Attn}(z_t, T)
=
\mathrm{softmax}\!\left(
\frac{(Q z_t)(K T)^{\!\top}}{\sqrt{d}}
\right)
(V T),
\]
where \(Q,K,V\) are learned linear projections.  
This formulation lets the model dynamically select which conditioning stream to
attend to at each spatial location. In early layers, \(t_{\mathrm{img}}\)
dominates to preserve object identity, while deeper layers increasingly rely on
\(t_{\mathrm{light}}\) and \(t_{\mathrm{pbr}}\) to synthesize correct shading,
specularity, and global illumination changes.

\paragraph{Mask-weighted fusion.}
From Sec.~\ref{subsec:mask_detail}, a spatial mask
\(M_{\theta}\in[0,1]^{H\times W}\) identifies illumination-sensitive regions.
We therefore modulate cross-attention using:
\[
\begin{aligned}
\mathrm{Attn}^{\star}(z_t, T)
&= (1 - M_{\theta}) \,\odot\, 
    \mathrm{Attn}_{\mathrm{base}}(z_t, t_{\mathrm{img}}) \\
&\quad+\; 
   M_{\theta} \,\odot\, 
   \mathrm{Attn}_{\mathrm{full}}(z_t, T).
\end{aligned}
\]
so that stable regions rely primarily on appearance cues while illumination-
variant areas receive full conditioning. This mechanism prevents unnecessary
texture distortion and preserves material fidelity under large lighting edits.

Overall, the combination of appearance, lighting, and proxy tokens enables
LightCtrl to achieve fine-grained, disentangled relighting while maintaining
strong geometric and photometric consistency.

\section{Training Details}
\label{sec:training_details}

All experiments are conducted on 4$\times$NVIDIA H800 GPUs using mixed-precision (fp16) training. 
We optimize LightCtrl with the AdamW optimizer, using an initial learning rate of \(\mathbf{1\times10^{-4}}\), 
\(\beta_{1}{=}0.9\), \(\beta_{2}{=}0.999\), and a weight decay of \(1\times10^{-2}\). 
The batch size is set to 16 per GPU (64 total), and the model is trained for 400K diffusion steps using a constant-with-warmup schedule with 5K warm-up steps. 
We adopt a DDIM noise scheduler with 1000 timesteps and follow the standard $v$-prediction objective. 
Unless otherwise stated, all models are selected using the best validation performance on the held-out ScaLight split.

We also evaluated LightCtrl on one NVIDIA H800(CUDA 12.2) using \textbf{bf16 precision} optimization. Results demonstrate high efficiency: the model ($\sim$1.2B parameters) generates a $512\times512$ image in just \textbf{$\sim$0.84s} (50 steps) with a peak memory usage of \textbf{2.53GB}.

\section{Implementation Details of the DPO Mechanism}
\label{sec:supp_dpo_details}

Our pipeline initializes with joint training using large-scale relighting pairs without ground-truth (GT) intrinsic supervision. While this implicit learning establishes strong generative priors, it can occasionally entangle material and lighting representations. To robustify the latent proxy encoder, we introduce a post-training stage using sparse PBR supervision, where only $\sim$3\% of the training data contains explicit annotations. Under such extreme label sparsity, we find that standard Supervised Fine-Tuning (SFT) suffers from severe generalization issues. Specifically, SFT merely demonstrates the ``correct'' state and often leads the model to overfit, causing it to memorize lighting artifacts such as baking cast shadows into the predicted albedo. 

To overcome this, we adopt Direct Preference Optimization (DPO) over SFT. DPO leverages a contrastive signal, teaching the model not just what constitutes valid physical consistency, but crucially, what destroys it. Implementation-wise, we construct preference pairs dynamically during training. We define the preferred positive sample $y_{pos}$ using the GT PBR maps. For the rejected negative sample $y_{neg}$, we use the model's own current prediction (i.e., $y_{neg} = E_\phi(x_s^{\ell_s})$). Because $y_{neg}$ is generated by the network itself, it inherently contains the specific artifacts and entangled shading that the model is prone to making. By constructing pairs in this manner, the training process explicitly penalizes the exact failure modes the model naturally produces.

\section{Impact of PBR Supervision Scale}
\label{subsec:pbr_scale}

To quantify how the amount of PBR-supervised data influences the quality of our
latent proxy and overall relighting performance, we conduct a controlled study
using different sizes of PBR annotations while keeping all other training
settings fixed. Specifically, we train LightCtrl with \{1K, 3K, 6K, 9K\}
PBR-supervised samples randomly drawn from ScaLight, and evaluate on the same
held-out object-level test set. Another 3K samples is collected in the Objaverse datasets.

Since the PBR encoder is the primary source of material and geometry priors, 
increasing the number of PBR-supervised examples naturally improves the fidelity 
of shading transfer and relighting accuracy. However, as shown in 
Table~\ref{tab:pbr_scale}, the gains diminish beyond roughly 9K supervised 
samples, indicating clear marginal returns. This saturation suggests that 
dense intrinsic annotations are unnecessary: even relatively sparse supervision 
is sufficient to provide stable material cues for the diffusion model and yield 
high-quality, controllable relighting.
\begin{table}[t]
\centering
\caption{
Effect of PBR supervision scale on relighting quality.
Increasing PBR annotations steadily improves performance, yet our experiments 
show that even sparse supervision provides meaningful benefits for relighting 
tasks.
}
\label{tab:pbr_scale}
\vspace{0.3em}
\resizebox{0.95\linewidth}{!}{
\begin{tabular}{lccc}
\toprule
\textbf{\#PBR Samples} & \textbf{RMSE} $\downarrow$ & \textbf{SSIM} $\uparrow$ & \textbf{PSNR} $\uparrow$ \\
\midrule
1K   & 0.468 & 0.487 & 12.8 \\
3K   & 0.421 & 0.534 & 14.7 \\
6K   & 0.331 & 0.612 & 19.3 \\
9K   & 0.194 & 0.873 & 23.5 \\
12K  & 0.105 & 0.921 & 28.6 \\
\bottomrule
\end{tabular}
}
\end{table}

\section{Additional Qualitative Results}

\begin{figure*}[t]
    \centering
    \includegraphics[width=\linewidth,trim=0 20 0 20,clip]{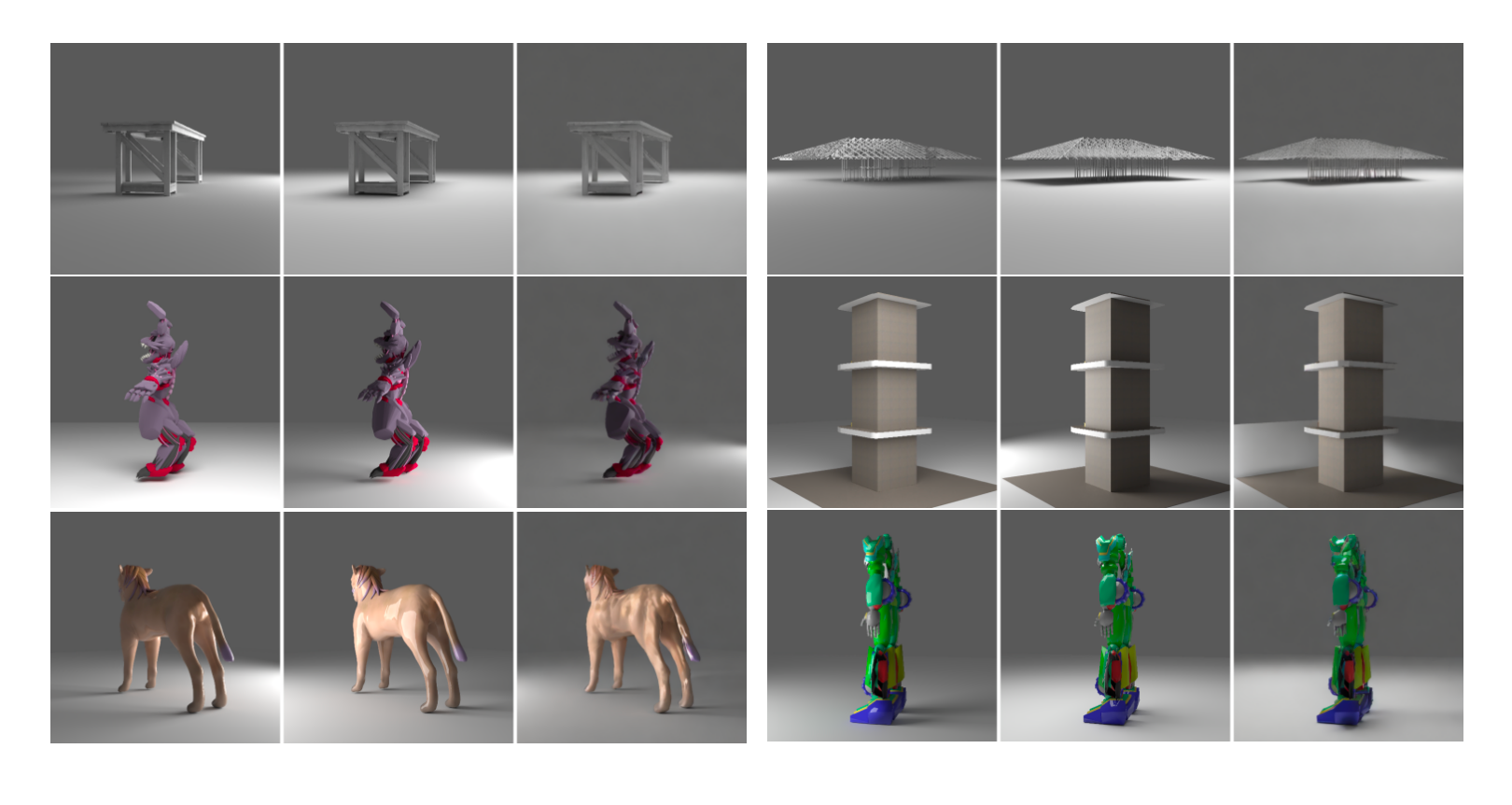}
    \caption{    Qualitative relighting results on \textbf{ScaLight} under varying \textbf{light positions}.
    Left: reference image; middle: ground-truth relit image; right: \textbf{LightCtrl} prediction.}
    \label{fig:pos1}
    \vspace{-5pt}
\end{figure*}

\begin{figure*}[t]
    \centering
    \includegraphics[width=\linewidth,trim=0 20 0 20,clip]{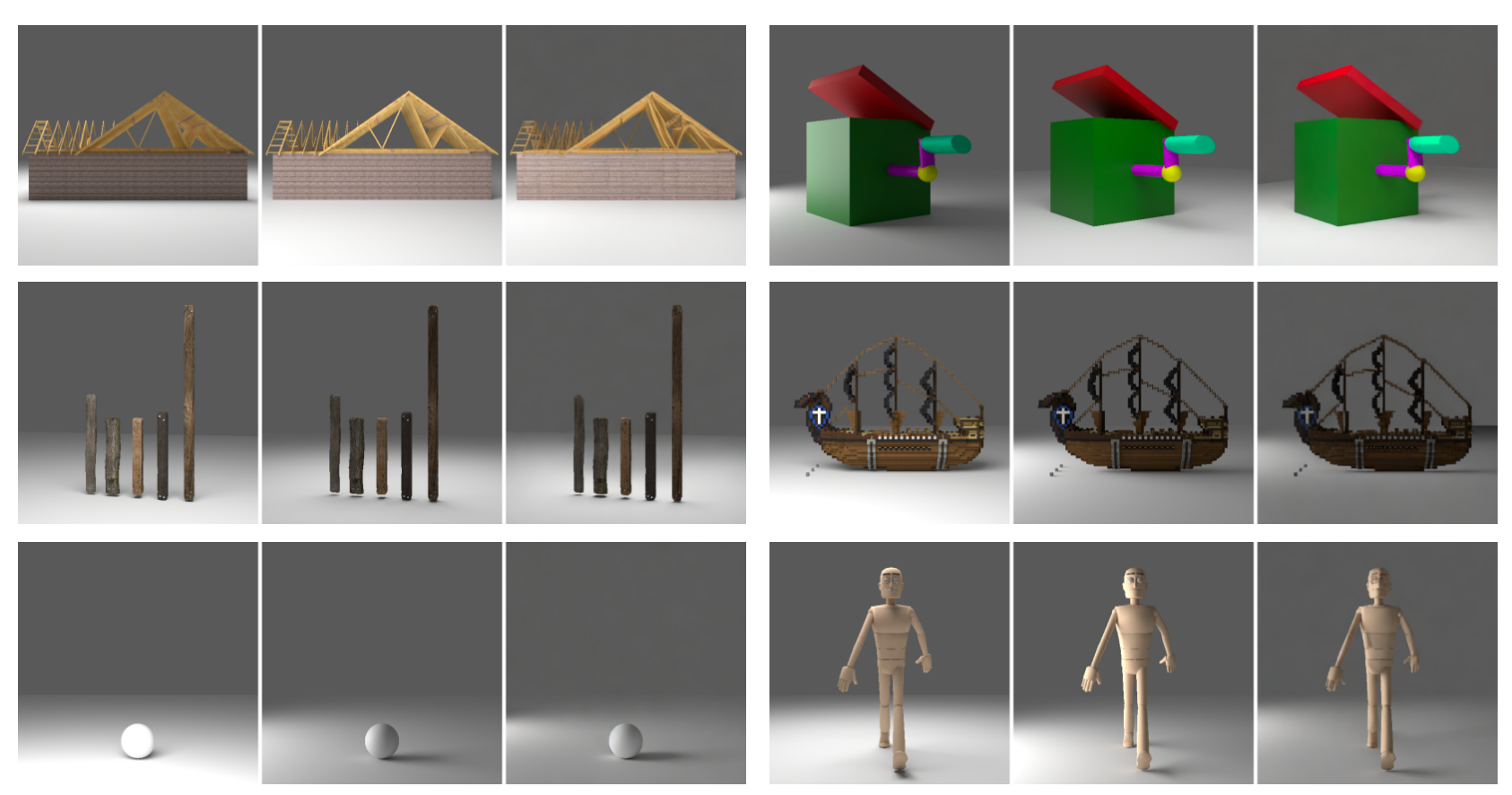}
    \caption{Qualitative relighting results on \textbf{ScaLight} under varying \textbf{light positions}.}
    \label{fig:pos2}
    \vspace{-5pt}
\end{figure*}

\begin{figure*}[t]
    \centering
    \includegraphics[width=\linewidth,trim=0 20 0 20,clip]{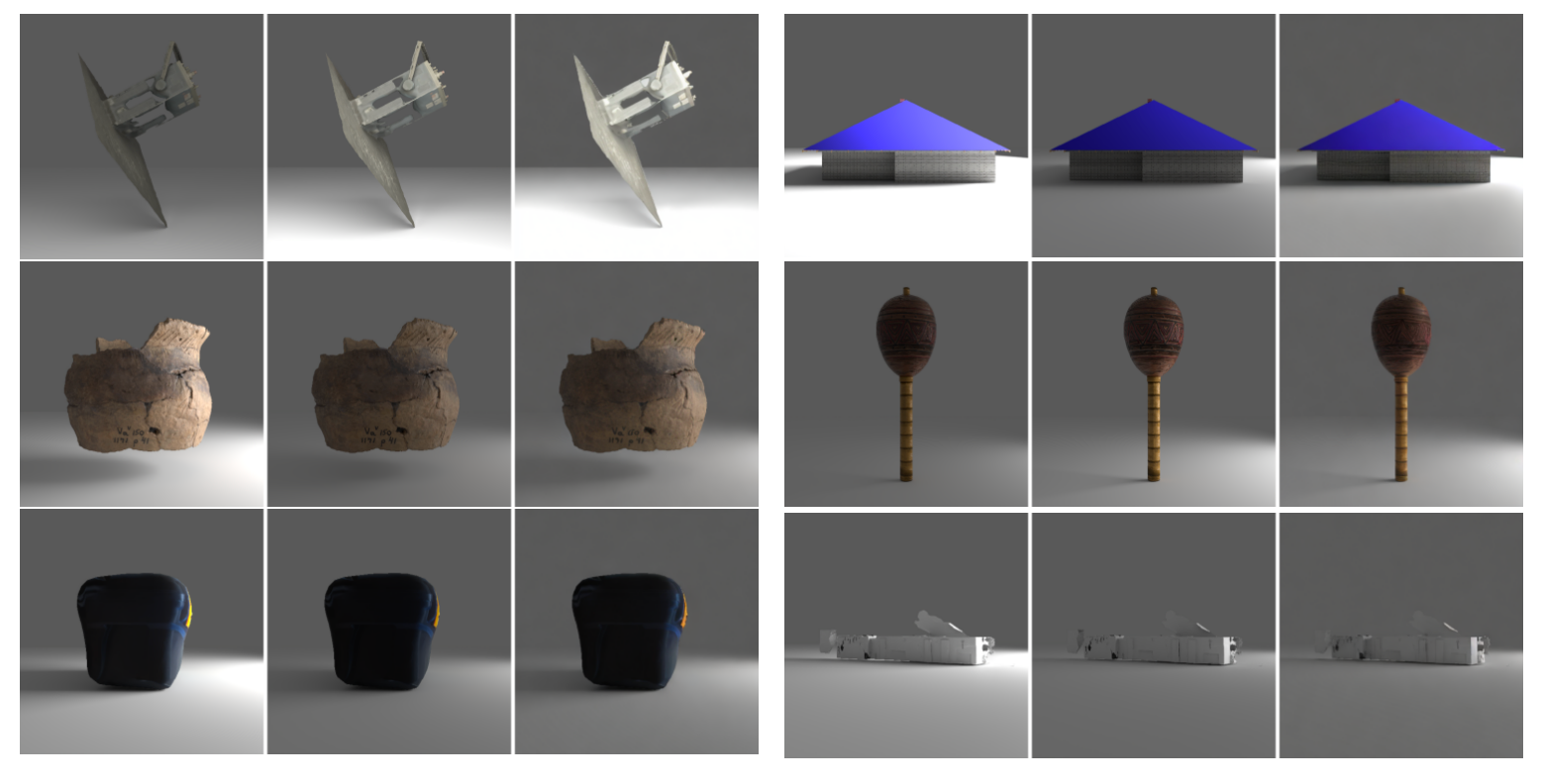}
    \caption{Qualitative relighting results on \textbf{ScaLight} under varying \textbf{light intensity}.
    Left: reference image; middle: ground-truth relit image; right: \textbf{LightCtrl} prediction.}
    \label{fig:energy1}
    \vspace{-5pt}
\end{figure*}

\begin{figure*}[t]
    \centering
    \includegraphics[width=\linewidth,trim=0 20 0 20,clip]{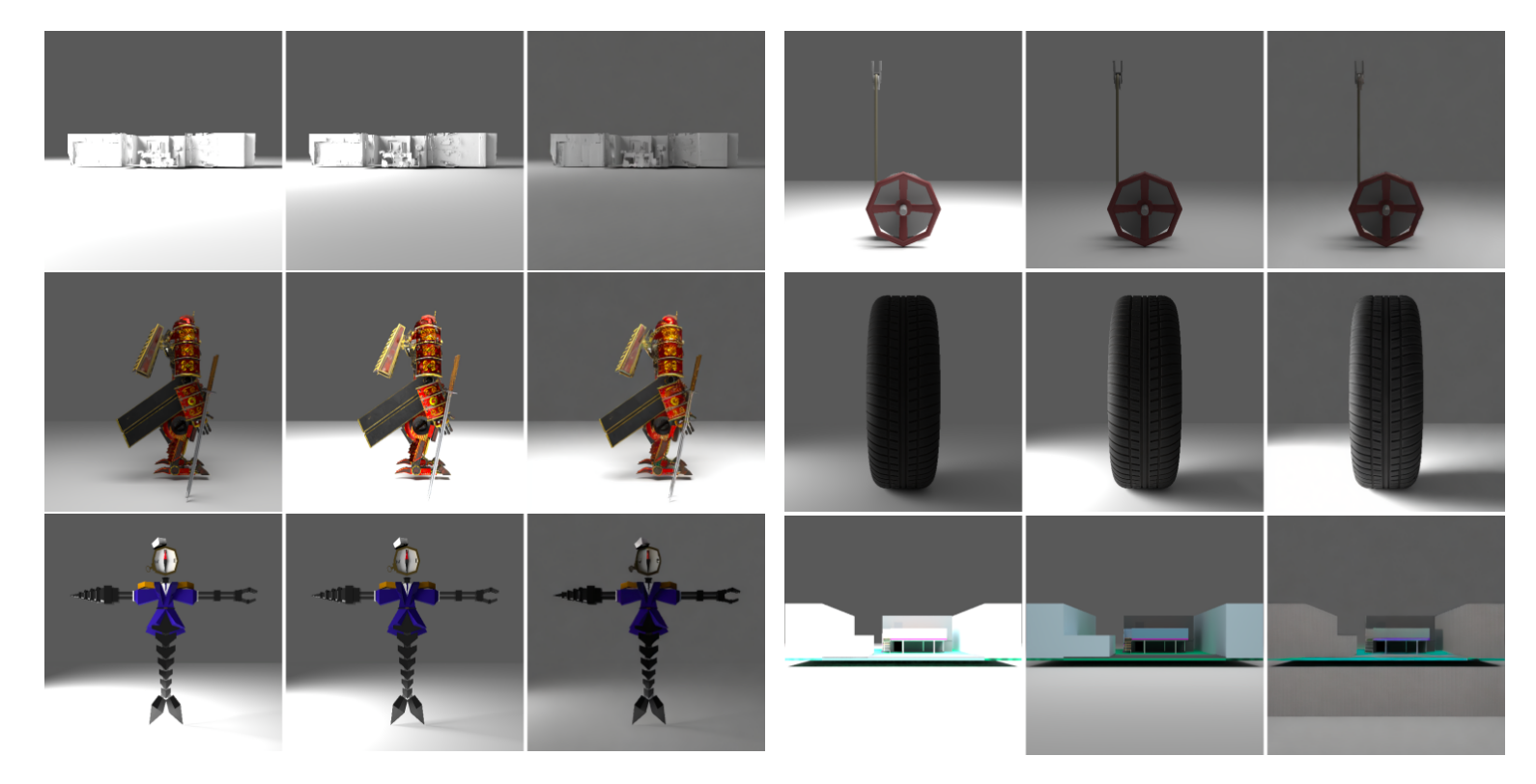}
    \caption{Qualitative relighting results on \textbf{ScaLight} under varying \textbf{light intensity}.}
    \label{fig:energy2}
    \vspace{-5pt}
\end{figure*}

\begin{figure*}[t]
    \centering
    \includegraphics[width=\linewidth,trim=0 20 0 20,clip]{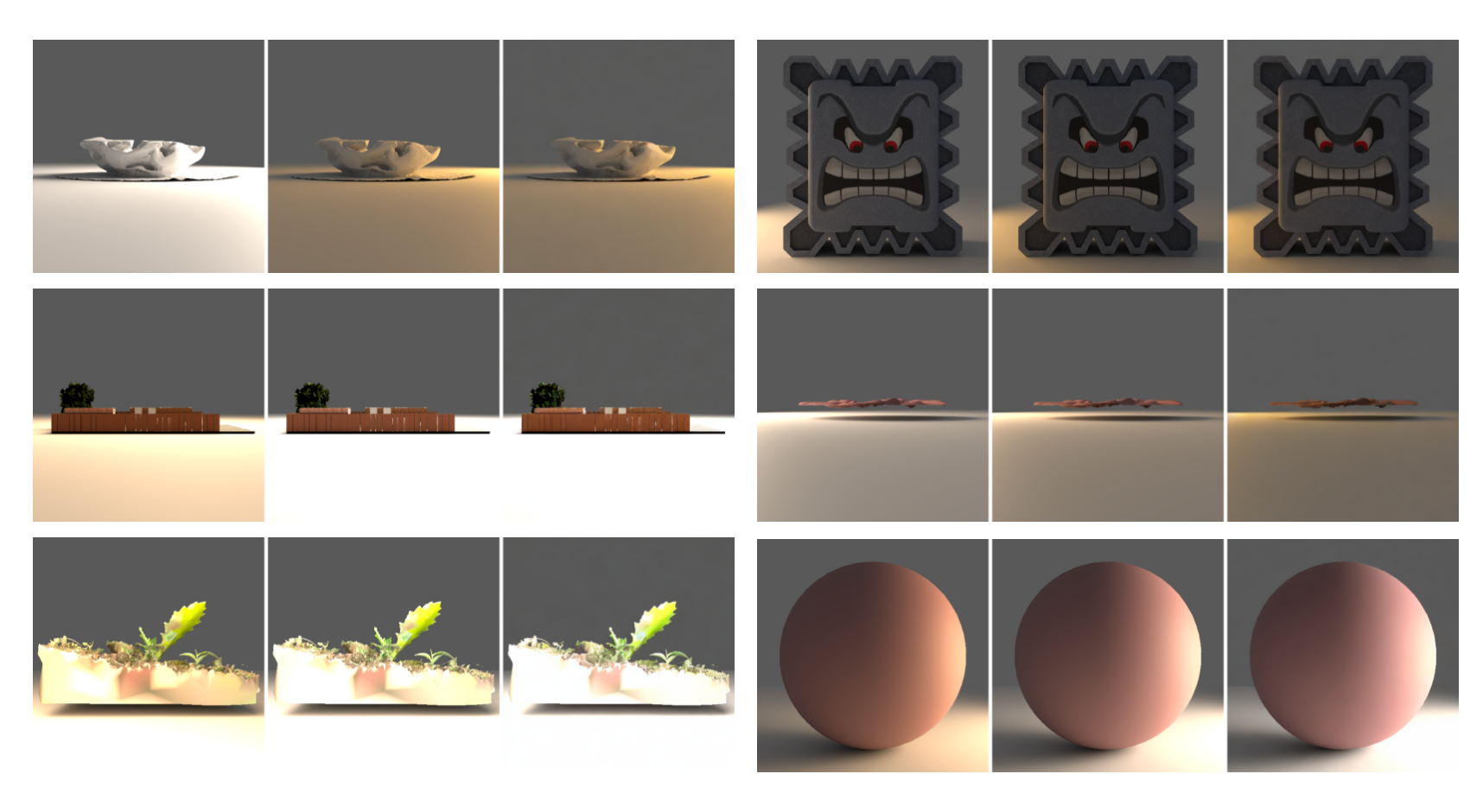}
    \caption{Qualitative relighting results on \textbf{ScaLight} under varying \textbf{light temperature}.
    Left: reference image; middle: ground-truth relit image; right: \textbf{LightCtrl} prediction.}
    \label{fig:color1}
    \vspace{-5pt}
\end{figure*}

\begin{figure*}[t]
    \centering
    \includegraphics[width=\linewidth,trim=0 20 0 20,clip]{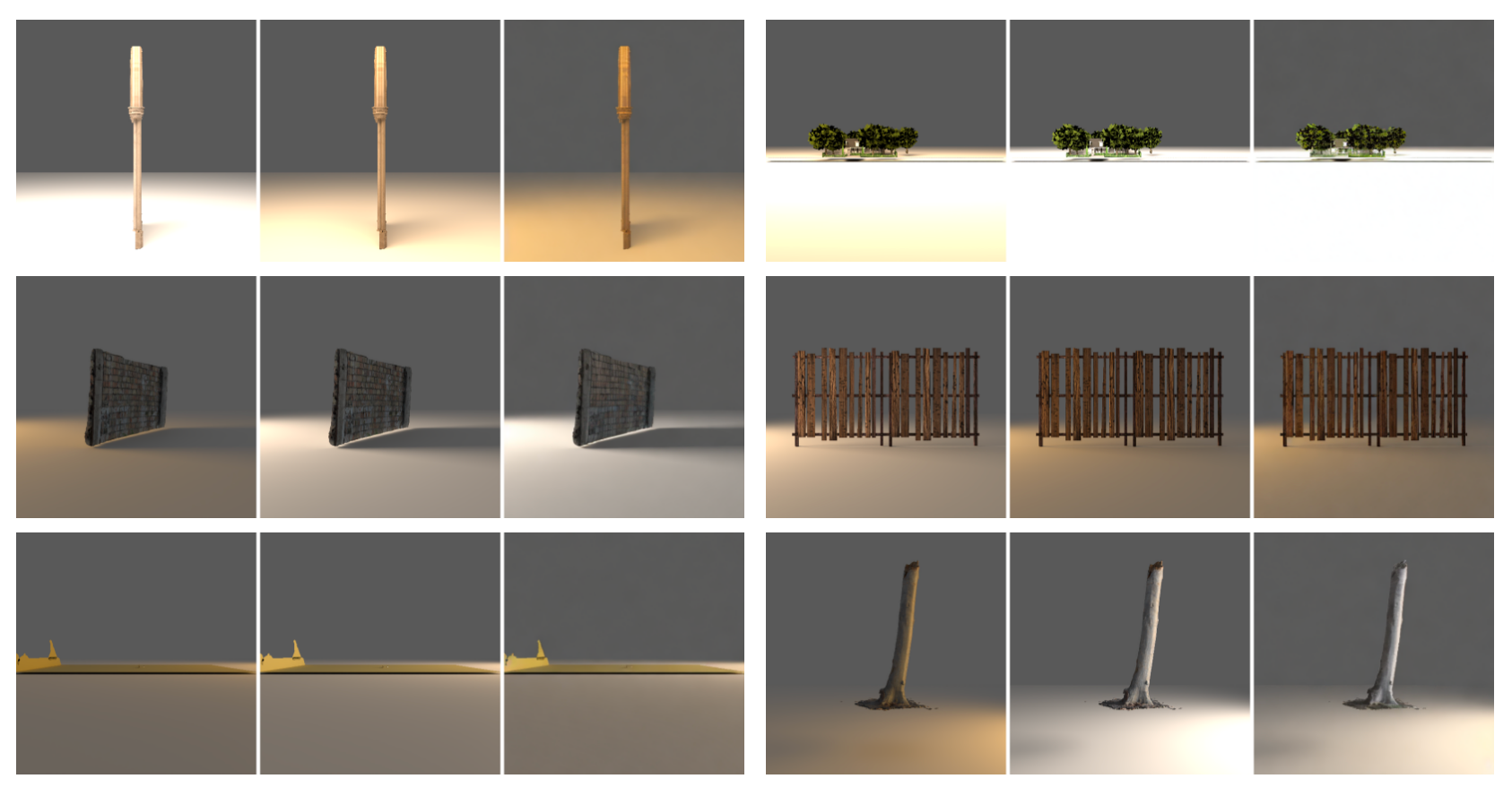}
    \caption{Qualitative relighting results on \textbf{ScaLight} under varying \textbf{light temperature}}
    \label{fig:color2}
    \vspace{-5pt}
\end{figure*}

\begin{figure*}[t]
    \centering
    \includegraphics[width=0.8\linewidth,trim=0 20 0 20,clip]{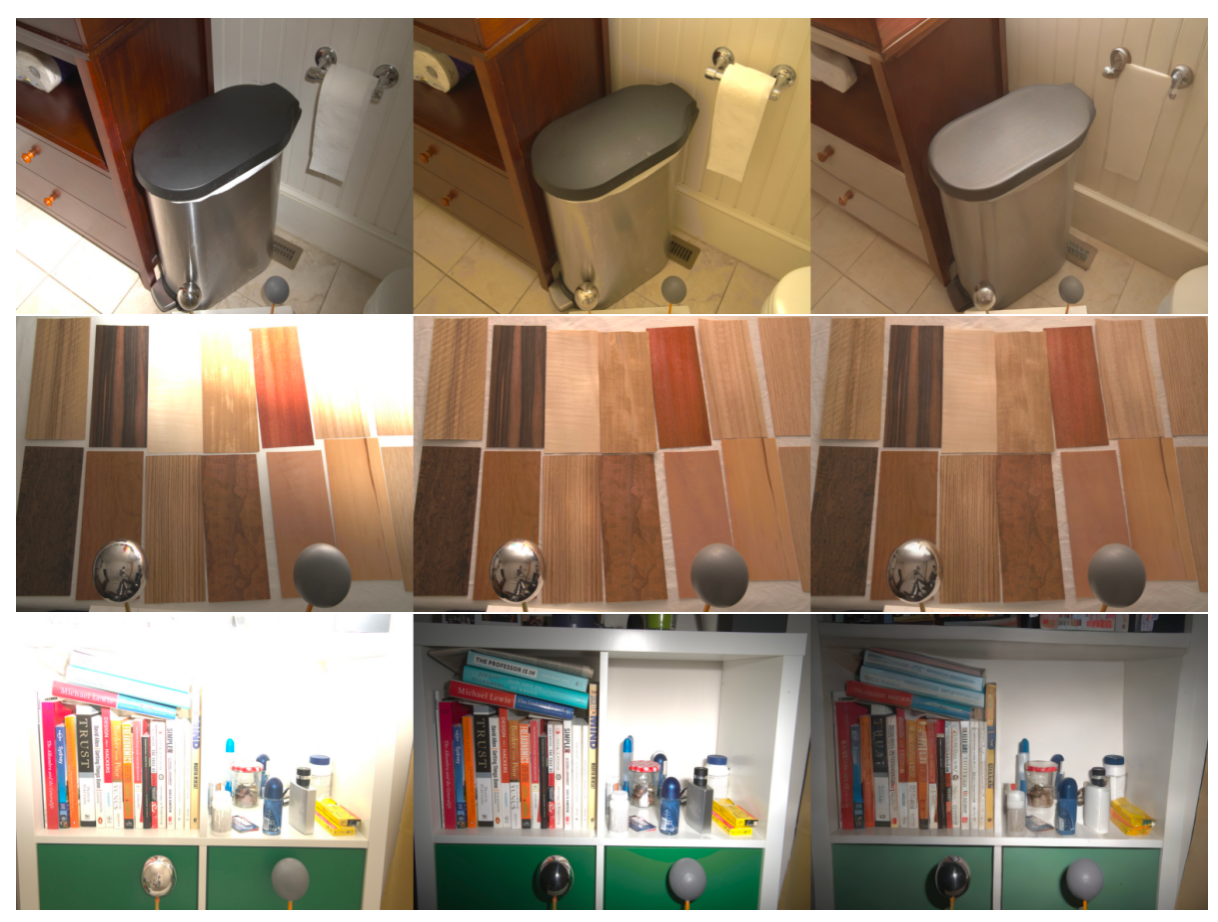}
    \caption{Qualitative relighting results on \textbf{MIIW}.
    Left: reference image; middle: ground-truth relit image; right: \textbf{LightCtrl} prediction.}
    \label{fig:scene1}
    \vspace{-5pt}
\end{figure*}

\begin{figure*}[t]
    \centering
    \includegraphics[width=0.8\linewidth,trim=0 20 0 20,clip]{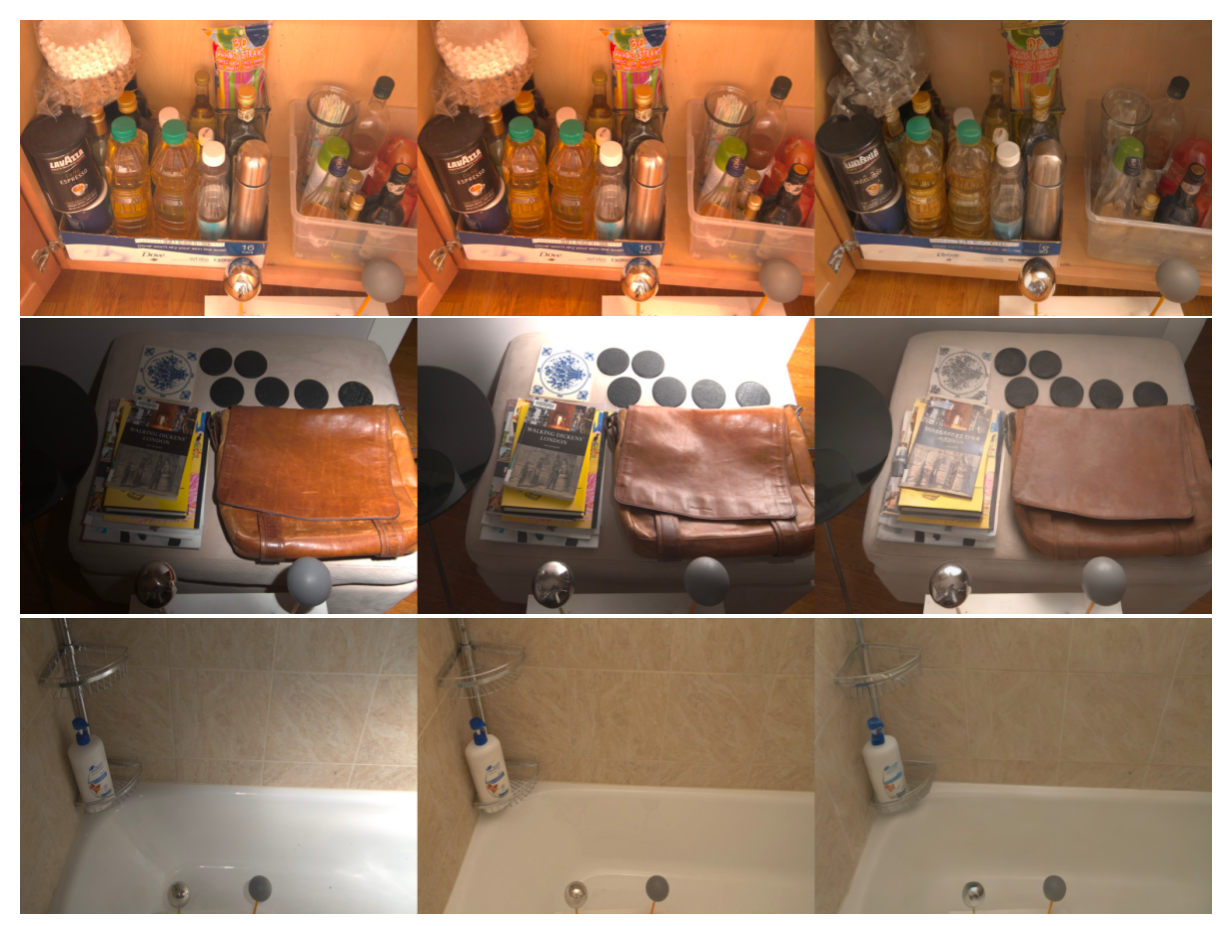}
    \caption{Qualitative relighting results on \textbf{MIIW}.
    Left: reference image; middle: ground-truth relit image; right: \textbf{LightCtrl} prediction.}
    \label{fig:scene2}
    \vspace{-5pt}
\end{figure*}

For completeness, we provide additional qualitative examples that complement
the results shown in the main paper. Figures~\ref{fig:pos1}, ~\ref{fig:pos2}, ~\ref{fig:energy1}, ~\ref{fig:energy2},  ~\ref{fig:color1}, ~\ref{fig:color2},
~\ref{fig:scene1} and ~\ref{fig:scene2} illustrate more relighting outcomes under controlled edits
of (i) light position (yaw--pitch), (ii) illumination intensity, and (iii)
color temperature. These samples demonstrate that LightCtrl produces smooth,
photometrically consistent transitions across continuous lighting variations. Finally, additional figures present additional scene-level results
on both synthetic environments and real-world images from MIIW.

\section{Conclusion}

In this supplementary document, we provided additional technical details of LightCtrl, including the formulation of our relative illumination encoding, the construction and training of the lighting-aware mask, and the integration of proxy, appearance, and lighting tokens within the conditioning architecture. We further described implementation specifics such as training schedules and hardware, and presented additional qualitative examples covering light position, intensity, color temperature, and scene-level relighting. Together, these materials complement the main paper by offering deeper insight into the design choices behind LightCtrl and by demonstrating its robustness under a wide range of controlled illumination conditions.

\newpage

\end{document}